\documentclass[sigconf]{acmart} 
\AtBeginDocument{%
  }
\theoremstyle{definition}

\setcopyright{acmlicensed}
\copyrightyear{2026}
\acmYear{2026}
\acmDOI{XXX}
\acmConference[Conference acronym 'XX]{The 2026 SIGKDD Conference}{April 13-17, 2026}{Jeju, Korea}
\acmISBN{XXX-X-XXXX-XXXX-X/2026/04}
\usepackage{bbm}
\usepackage{graphicx}   
\usepackage{subcaption} 
\usepackage{threeparttable}
\usepackage{soul}

\usepackage{enumitem}
\begin{document}

\title[GCN-MPPR: Enhancing the Propagation of MPNNs via Motif-Based Personalized PageRank]{GCN-MPPR: Enhancing the Propagation of Message Passing Neural Networks via Motif-Based Personalized PageRank}

\author{Mingcan Wang}	
\email{mingcan\_wl@163.com}
\affiliation{%
  \institution{School of Computer Science and Engineering, Northeastern University}
  \city{Shenyang}
  \state{Liaoning}
  \country{China}
}

\author{Junchang Xin*}
\email{xinjunchang@mail.neu.edu.cn}
\affiliation{%
  \institution{School of Computer Science and Engineering, Northeastern University}
  \institution{Key Laboratory of Big Data Management and Analytics (Liaoning Province), Northeastern University}
  \city{Shenyang}
  \state{Liaoning}
  \country{China}
}

\author{Zhongming Yao}
\email{zyao@cs.aau.dk}
\affiliation{%
  \institution{The Department of Computer Science, Aalborg University}
  \city{Aalborg}
  \country{Danmark}
}

\author{Xian Zhang}
\affiliation{%
  \institution{School of Computer Science and Engineering, Northeastern University}
  \city{Shenyang}
  \state{Liaoning}
  \country{China}
}

\author{Kaifu Long}
\affiliation{%
  \institution{School of Computer Science and Engineering, Northeastern University}
  \city{Shenyang}
  \state{Liaoning}
  \country{China}
}

\author{Zhiqiong Wang}
\email{wangzq@bmie.neu.edu.cn}
\affiliation{%
\institution{College of Medicine and Biological Information Engineering, Northeastern University}
\city{Shenyang}
\state{Liaoning}
\country{China}}
\renewcommand{\shortauthors}{Wang et al.}

\begin{abstract}

The algorithms based on message passing neural networks (MPNNs) on graphs have recently achieved great success for various graph applications. However, studies find that these methods always propagate the information to very limited neighborhoods with shallow depth, particularly due to over-smoothing. That means most of the existing MPNNs fail to be so `deep'. Although some previous work tended to handle this challenge via optimization- or structure-level remedies, the overall performance of GCNs still suffers from limited accuracy, poor stability, and unaffordable computational cost. Moreover, neglect of higher-order relationships during the propagation of MPNNs has further limited the performance of them. 

To overcome these challenges, a novel variant of PageRank named motif-based personalized PageRank (MPPR) is proposed to measure the influence of one node to another on the basis of considering higher-order motif relationships. Secondly, the MPPR is utilized to the message passing process of GCNs, thereby guiding the message passing process at a relatively `high' level. The experimental results show that the proposed method outperforms almost all of the baselines on accuracy, stability, and time consumption. Additionally, the proposed method can be considered as a component that can underpin almost all GCN tasks, with DGCRL being demonstrated in the experiment. The \textbf{anonymous} code repository is available at: https://anonymous.4open.science/r/GCN-MPPR-AFD6/. 
\end{abstract}

\maketitle

\section{Introduction}\label{intro}
Graph is a ubiquitous structure that garners significant interest from both academic and industrial communities \cite{graph3, graph5, graph11, graph1, Zhao}. Effective mining on graphs will benefit a wide range of applications, such as web advertising \cite{wa} and social network analysis \cite{sna}. The deep learning methods have achieved remarkable successes in different analysis levels of node, link and graph, which cannot be neglected in any possible manner. There are many approaches for conducting deep learning algorithms on graphs. Early methods such as node embedding techniques \cite{NE01, NE02} typically rely on random walks or matrix factorization to learn node representations directly, often in an unsupervised fashion and without incorporating node features. In contrast, many modern approaches jointly leverage both graph structure and node features within a supervised (or semi-supervised) framework. Representative examples include spectral graph convolutional neural networks \cite{SGCNN1, SGCNN2, SGCNN3}, message passing (neighbor aggregation) algorithms \cite{MP1, MP2, MP3}, and neighbor aggregation methods based on recurrent neural networks \cite{RNN1, RNN2, RNN3}. 


Among these methods, message passing neural networks (MPNNs) have received particularly intense attention in recent years for its flexibility and good performance. The past few years have seen significant advances in enhancing the message passing mechanism, including attention-based approaches \cite{attention1, attention3}, stochastic process integrations \cite{stochastic1, stochastic3}, and so on. Despite these impressive developments, the robustness and scalability of message passing algorithms remain frequently questioned. In MPNNs, sufficient layers of propagation are necessary to capture more information of nodes within the network, as deeper architectures allow aggregation of multi-hop neighborhood information and thus model longer-range dependencies more effectively. However, aggregation through averaging often results in highly similar learned features, a phenomenon widely known as over-smoothing. As a result, existing message-passing neural networks are typically constrained by shallow-layer information, limiting their ability to fully exploit the rich structural patterns in large or complex graphs.


Recent study \cite{ppnp} proposed to combine personalized PageRank (PPR) with the message passing neural network, and came up with a novel framework that emphasizes `predict and propagate' to solve the problem. Since PPR uses the idea of random walk, the proposed method can propagate the information more deeply without adding layers with the aid of PPR. And their work has broadened the development of the message passing algorithms \cite{PPRGo, Zebra}. However, all of these methods are about MPNNs plus a linear model (e.g. PPR), neglecting the higher-order relationships. This prevents the proposed method from capturing the higher-order relationships and guiding the propagation based on them. Consequently, the current propagations of MPNNs are still restricted by limited shallow information. And the following challenges faced by existing works restricts the performance of MPNNs. 


\textbf{Challenge I: Depth-induced homophily collapse. } Despite ongoing efforts to mitigate over-smoothing through techniques like structure-level \cite{sl1, sl3} or optimization-level \cite{ol1, ol3, ol5} remedies, the over-smoothing problem has not been completely resolved. Shallow-layer propagation would result in shallow information of nodes, while simply enlarging the number of layers in MPNNs cannot effectively enable the model to acquire more discriminative information and may even result in homophily collapse. As a result, developing more effective propagation methods remains one of the major challenges. 

\textbf{Challenge II: Failure to utilize higher-order relationships. } In the predominant frameworks of MPNNs as well as many variants of them such as VGAE\cite{VGAE} and PPNP \cite{ppnp}, the vast majority of models rely heavily on linear transformations and linear gradient descent throughout their message-passing and updating steps (e.g., linear aggregations and linear projections). This linearity-centric design dominates the methods and largely overlooks the prevalence of higher-order relational structures — such as motifs — that are ubiquitous in real-world networks and play a crucial role in network science. Consequently, conventional MPNNs struggle to capture rich, non-pairwise dependencies and multi-way interactions, limiting their expressive power on heterophilic, hierarchical, or combinatorially complex graphs.

\textbf{Challenge III: Trade-offs in accuracy, stability, and computational efficiency.} In MPNNs, accuracy, stability, and computational efficiency are all critically important; a significant compromise in any one of these dimensions will directly impair the model's predictive performance, training reliability, and real-world scalability. Existing remedies—whether structure-level modifications  or optimization techniques—often improve one aspect at the expense of others, leading to unstable training, high computational overhead, or marginal accuracy gains across tasks like node classification and link prediction. Then how to modify the MPNN models to ensure an overall improvement is another challenge.

\begin{figure}[!t]
  \centering
  \includegraphics[width=0.95\linewidth]{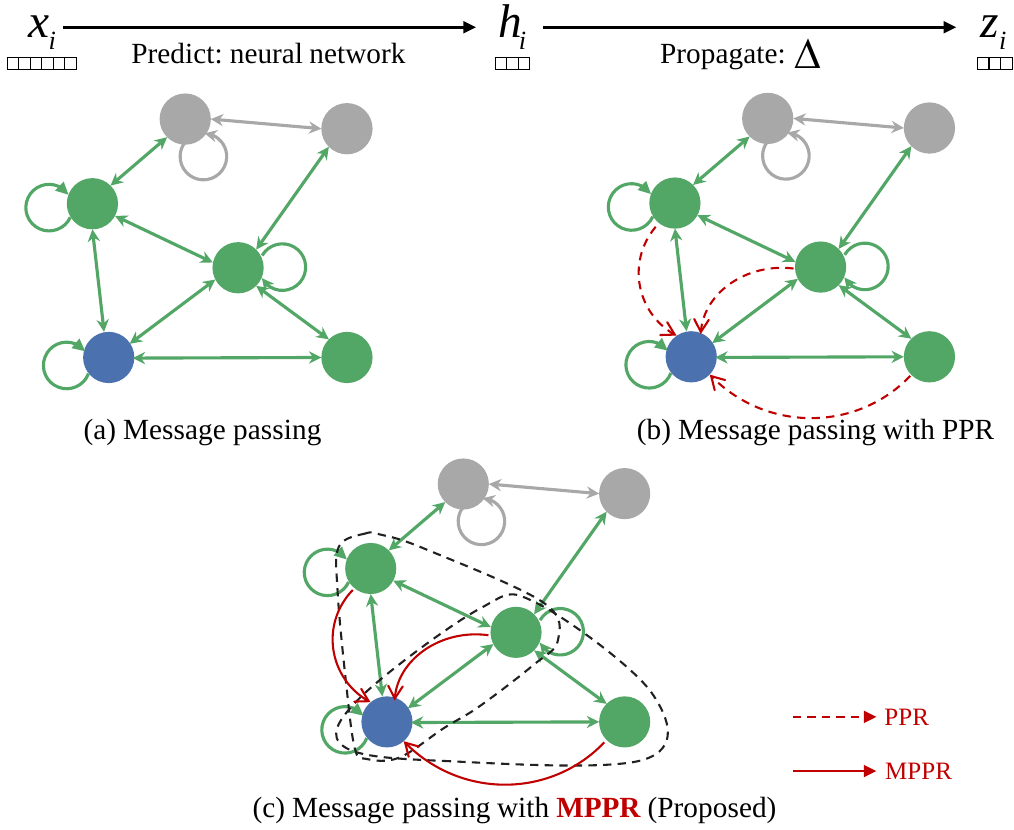}
  \caption{Visualization of the contributions of GCN-MPPR upon existing works, whereas $\Delta$ = (a) message passing; (b) message passing with PPR and (c) message passing with MPPR (spreading information further considering higher-order motifs). \label{fig-intro}}
\end{figure}



To alleviate the problems mentioned above, firstly, a novel network metric named motif-based personalized PageRank (MPPR) is proposed to measure and reflect the influence of one node to another on the basis of considering higher-order motif relationships. Secondly, the MPPR is utilized to the message passing process of GCNs, yielding a novel framework named GCN-MPPR, which tends to guide the propagation of message passing process based on MPPR. This would allow enlarging the covering domain of specific nodes without adding the layers of GCNs and causing over-smoothing. The differences between GCN-MPPR and existing work are visualized in \autoref{fig-intro}. Theoretically, since MPPR emphasize the higher-order relationships in the graphs, message passing via MPPR can guide the training process more accurately and efficiently on the basis of considering higher-order motif relationships, thus upgrading the accuracy, stability and computational efficiency. Experimentally, it is demonstrated that GCN-MPPR is competitive among the baselines in both node classification and link prediction tasks. Furthermore, experiments also verify the GCN-MPPR can be easily integrated into many state-of-the-art methods as a component and upgrade the performance. To summarize, the contributions of this work can be summarized as:  

\begin{itemize}[leftmargin=*]

\item {Proposing MPPR, a novel network metrics to measure the influence of one node to another based on PageRank and motif, that can inherit the expressive power of motifs to effectively capture complex, higher-order relationships beyond traditional pairwise interactions. }
\item{Applying MPPR to the propagation of MPNNs, which enables a more accurate and effective message aggregation process in different tasks of graphs, enabling the models to be more faithful for higher-order relational patterns without incurring prohibitive computational overhead. }
\item{Conducting thorough experiments, demonstrating that the proposed method not only achieves competitive performance against competitors across various tasks, but also serves as a plug-and-play component that consistently improves performance when integrated into MPNN models. }
\end{itemize}

The rest of the paper is organized as follows. Related work is reviewed in \autoref{sec2}. Section \ref{sec3} defines the problem and the proposed methods.
The proposed method and the competitors are evaluated in \autoref{sec4}. Finally, \autoref{sec5} concludes the paper. Relative appendixes appear in Appendix. 

\section{Related Work}
\label{sec2}
\subsection{Network Motif} Motif, also known as graphlets or subgraphs, characterizes higher-order relations in complex networks, which was first introduced in \cite{MotifFirst}. Rather than consider the linear relationship within each network only, motif can capture the multi-node interactions in the topological structure. For example, in social networks, the existence of the `triangular motif' (A $\leftrightarrow$ B, A $\leftrightarrow$ C, B $\leftrightarrow$ C) can enhance the cohesion of the group. The influence of A on B is indirectly amplified through C (the interaction between B and C is jointly influenced by A), which is a common nonlinear relationship. Then the edge existing in more triangular motifs is considered more important than the one existing in less or none triangular motifs. Also in web graphs, the more motifs a link is evolved, the more important the link should be considered. Therefore, the capacity of capturing nonlinear relationships makes motifs useful in many applications, such as web networks, social networks, neuroscience networks and bioinformatics networks.

As most of the previous work focused on how to efficiently count the number of motifs in complex networks \cite{count1, count2, count3, count4}, it was proven that motifs can be used for graph/web mining tasks \cite{motifweb1, motifweb2}. For instance, Zhao {\it et al.} \cite{MPR} proposed to rank the users in social networks via motif-based PageRank, which enhances the ranking ability of PageRank by capturing higher-order relationships using motif and incorporating them into PageRank. Wu {\it et al.} \cite{motif2} introduced a motif-based contrastive graph clustering approach with clustering oriented prompt, which employs a specialized Siamese encoder network to obtain both lower-order and higher-order node embeddings. The encoder processes both views of the graph: one based on lower-order adjacency and the other on higher-order motif structures, and the higher-order motif (such as triangles) is extracted using motif adjacency matrices. 

\subsection{PageRank and Personalized PageRank}

Apart from ranking web pages as its original aim,  PageRank (PR) \cite{PageRank} has been used in domains as vast as web graph analysis, resource allocation and so on. For example, Zhao {\it et al.} \cite{MPR} proposed to rank the social network users via motif-based PageRank, a variant of PageRank that emphasize higher-order relationships. Zhao {\it et al.} \cite{MPRZhaohai} proposed to rank the Chinese medicine acupoints by introducing higher-order interactions between multiple acupoints and PageRank in the acupoint-disease network. As a variant of the PageRank, Personalized PageRank (PPR) is designed to measure the importance of nodes in a graph with respect to a specific `personalization' node (or set of nodes). In PageRank, each node's score is a weighted sum of the scores of its neighbors, with a small probability of teleporting to a specific node in the graph. Comparatively, PPR modifies this by restricting teleportation to a specific set of personalization node(s), thus biasing the scores toward nodes that are close to these personalization node(s) in the graph. This makes it more than popular in many applications, such as recommendation systems \cite{RS1, RS3}. 

\subsection{MPNN Meets Personalized PageRank} 

In order to alleviate the over-smoothing problem of normal MPNNs, Gasteiger {\it et al.} \cite{ppnp} proposed to use the relationship between graph convolutional networks (GCN) and PageRank to derive an improved propagation scheme based on personalized PageRank, namely PPNP. And they proposed a approximate version of it, named approximate PPNP (APPNP). Inspired by this, Bojchevski {\it et al.} \cite{PPRGo} proposed to scale GNN upon networks with millions of nodes by utilizing an efficient approximation of information diffusion, resulting in significant speed gains while maintaining relatively state-of-the-art prediction performance. Li {\it et al.} \cite{Zebra} built the theoretical link between the temporal message passing scheme adopted by T-GNNs and the temporal random walk process on dynamic graphs, which indicates that it would be possible to select a few influential temporal neighbors to compute a target node’s representation without compromising the predictive performance. They then proposed to utilize T-PPR, a parameterized metric for estimating the influence score of nodes on evolving graphs.  Ma {\it et al.} \cite{PPRTGI} proposed personalized PageRank graph neural network for TF-target gene interaction detection method, namely PPRTGI. 

\subsection{Main Contributions against Related Works}

Compared to the previous works, first and foremost, Motif-Based Personalized PageRank (MPPR) is proposed, a latest variant of PPR or PR. MPPR can rank the importance between a pair of nodes on the basis of considering the high-order motifs of the network structure, thereby being more receivable and comprehensive. Secondly, inspired by PPNP \cite{ppnp}, MPPR is applied to the message passing process of graph convolutional networks, yielding GCN-MPPR, enabling the information of nodes to spread farther and faster without enlarging the number of layers and causing over-smoothing, thus improving the accuracy and computational efficiency of GCN models. Finally, the proposed MPPR can be regarded as a component, which can play an active role in speeding up and improving the accuracy in many MPNN architectures, with DGCRL being verified as an example in this paper. 

\section{Graph Neural Networks with Motif-Based Personalized PageRank} 
\label{sec3}

In this section, we first introduce the notations and definitions of the paper. Second, the details of MPPR is discussed, which is a latest variant of PPR. Then, the GCN-MPPR framework for both node classification and link prediction is detailed. 

\subsection{Notations and Preliminaries}

Let $G = \left( V, E\right)$ be an unweighted directed graph, where $V = \left\{v_1, ..., v_n\right\}$ is node set with cardinality $|V| = n$ and $E = \left\{e_1, ..., e_m\right\}$ is edge set with cardinality $|E| = m$. The features of nodes are denoted as $\boldsymbol{X}\in {{\mathbb{R}}^{n\times f}}$, with the number of features notated as $f$ for each node. And the labels (or classes) are denoted as $\boldsymbol{Y}\in {{\{0, 1\}}^{n\times C}}$ with the number of classes denoted as $C$. The graph $G$ is described by the adjacency matrix $\boldsymbol{A}\in {{\mathbb{R}}^{n\times n}}$, while $\overset{\sim}{\boldsymbol{A}}\, = \boldsymbol{A}+{\boldsymbol{{I}}_{n}}$ denotes the adjacency matrix with added self-loops. Denote $\overset{\sim}{\mathop{\boldsymbol{A}}}\,' = \overset{\sim}{\boldsymbol{D}}\,^{-1/2}\, \overset{\sim}{\boldsymbol{A}}\, \overset{\sim}{\boldsymbol{D}}\,^{-1/2}\,$ as the  symmetrically normalized adjacency matrix with self-loops, whereas $\overset{\sim}{\boldsymbol{D}}\,_{ij} = \sum\nolimits_{k}{\overset{\sim}{\mathop{{{A}_{ik}}}}\,{{\delta }_{ij}}}$ is the diagonal degree matrix. Given the above notations, the PageRank value vector $\boldsymbol{ \psi}$ of the nodes can be calculated by iteration as,
\begin{equation}\label{eq1}
\boldsymbol{ \psi}_t = d\cdot\textbf{P}^T\boldsymbol{ \psi}_{t-1} + \frac{1-d}{n}\boldsymbol{e}
\end{equation}
where $d\in \left(0, 1\right]$ is the damping factor, $t$ is the iteration time, $\boldsymbol{e} \in {{\mathbb{R}}^n}$ is a vector with each element equal to 1 and $\textbf{P}$ is obtained by $\textbf{P} = A_{ij}/\sum\nolimits_{j}{{{A}_{ij}}}$. And the calculation process has been proved to be converged \cite{PageCon}. As the concept of personalized PageRank defines the root node $v_i$ via the teleport vector $i_{v_i}$, a one-hot indicator vector, the adaptation of personalized PageRank can be obtained for node $x$ using: ${{\pi }_{\text{ppr}}}\left( {v_i} \right)=\left( 1-\alpha  \right)\cdot \overset{\sim}{\mathop{\boldsymbol{A}}}\,'\cdot {{\pi }_{\text{ppr}}}\left( {v_i} \right)+\alpha \cdot {i_{v_i}}$, with the teleport (or restart) possibility $\alpha \in \left(0, 1\right]$. By solving the equation, we obtain ${{\pi }_{\text{ppr}}}\left( {{v_i}} \right)=\alpha \cdot {{\left( {{\boldsymbol{I}}_{n}}-\left( 1-\alpha  \right)\overset{\sim}{\mathop{\boldsymbol{A}}}\,' \right)}^{-1}}{i_{v_i}}$. Since $i_{v_i}$ is a vector containing only one `1' value at the corresponding position except `0', which allows us to preserve the node’s local neighborhood even in the limit distribution, it is possible to union ${\pi}_{\text{ppr}}(v_1)$, ..., ${\pi}_{\text{ppr}}(v_n)$, yielding the full personalized PageRank matrix: 
\begin{equation}\label{eq2}
{{\Pi }_{\text{ppr}}}=\alpha \cdot {{\left( {{\boldsymbol{I}}_{n}}-\left( 1-\alpha  \right)\overset{\sim}{\mathop{\boldsymbol{A}}}\,' \right)}^{-1}}
\end{equation}

For the ease of description of the following contents, the definitions of motif and motif set is borrowed from \cite{MotifDef, MPR} and presented.

\begin{definition} \label{def1} Motif. A motif $M$ is defined on $k$ nodes by a tuple $\left(\boldsymbol{B}, \mathcal{A}\right)$, where $\boldsymbol{B}$ is a $k \times k$ binary matrix, and $ \mathcal{A} \in \{1, 2, ..., k\}$ specifies the anchor set, which is the set of the indices of the anchor nodes. \end{definition}

In $Definition$ {\it \ref{def1}}, a graph encoding the edge patterns between the $k$ nodes are represented by $\boldsymbol{B}$ and a subset of the $k$ nodes for defining the motif-based adjacency matrix is denoted as $\mathcal{A}$. To put it differently, two nodes will be regarded as occurring in a given motif only when their indices belong to $\mathcal{A}$. As usual, anchor nodes are all of the $k$ nodes, in which case the motif is called simple motif; Otherwise, it is called anchored motif. In this work, we only consider the case of simple motif, and more specifically three-node simple motif. Given the definition above, we can define the set of motif as follows. 

\begin{definition} \label{def2} Motif set. A motif set, notated as $\mathcal{M}\left(\boldsymbol{B}, \mathcal{A}\right)$, in an unweighted directed graph $G$ with adjacency matrix $\boldsymbol{A}$ is defined as:
\begin{align}\label{eqdef2}
&\mathcal{M}\left(\boldsymbol{B}, \mathcal{A}\right) \nonumber \\
 = &\left\{ \left( set\left( v \right),set\left( {{\chi }_{\mathcal{A}}}\left( v \right) \right) \right)|v\in {{V}^{k}},{{v}_{1}},...,{{v}_{k}},distinct,{{\boldsymbol{A}}_{\textbf{v}}}=\boldsymbol{B} \right\}
\end{align}
where $\textbf{v}$ is an ordered vector representing the indices of $k$ nodes, and ${\boldsymbol{A}}_{\textbf{v}}$ is the $k\times k$ adjacency matrix of the subgraph induced by \textbf{v}. ${\chi }_{\mathcal{A}}\left(\cdot \right)$ is a selection function that takes the subset of a $k$-tuple indexed by $\mathcal{A}$, and set$\left( \cdot \right)$ is an operator that transforms an ordered tuple to an unordered set, thus set$\left(\left(v_1, v_2, ..., v_k\right)\right) = \left\{v_1, v_2, ..., v_k \right\}$. 
\end{definition}

\begin{figure*}[!t]
    \centering
    \begin{subfigure}[t]{0.58\textwidth}
        \centering
        \includegraphics[width=\textwidth]{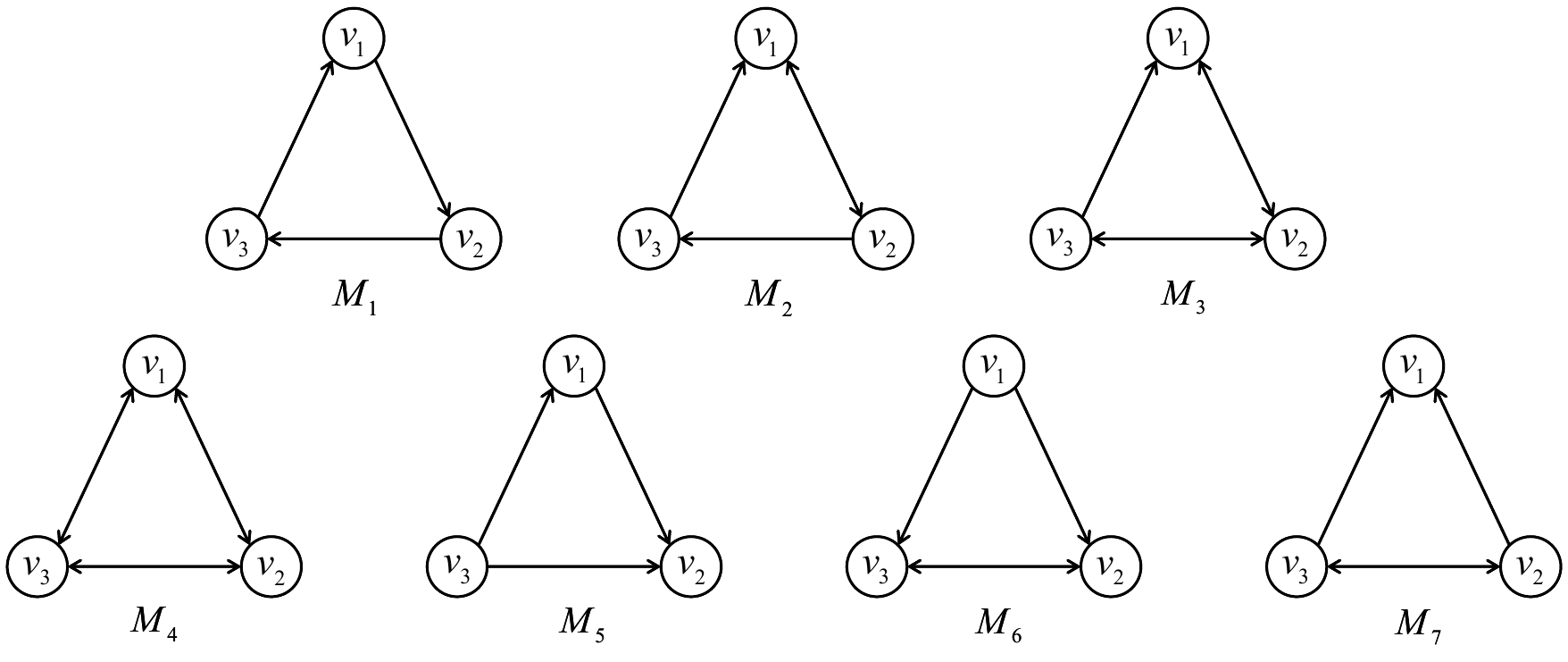}
        \caption{Seven Types of Triangle Motifs.}
        \label{fig:sub1}
    \end{subfigure}
    \hfill 
    \begin{subfigure}[t]{0.34\textwidth}
        \centering
        \includegraphics[width=\textwidth]{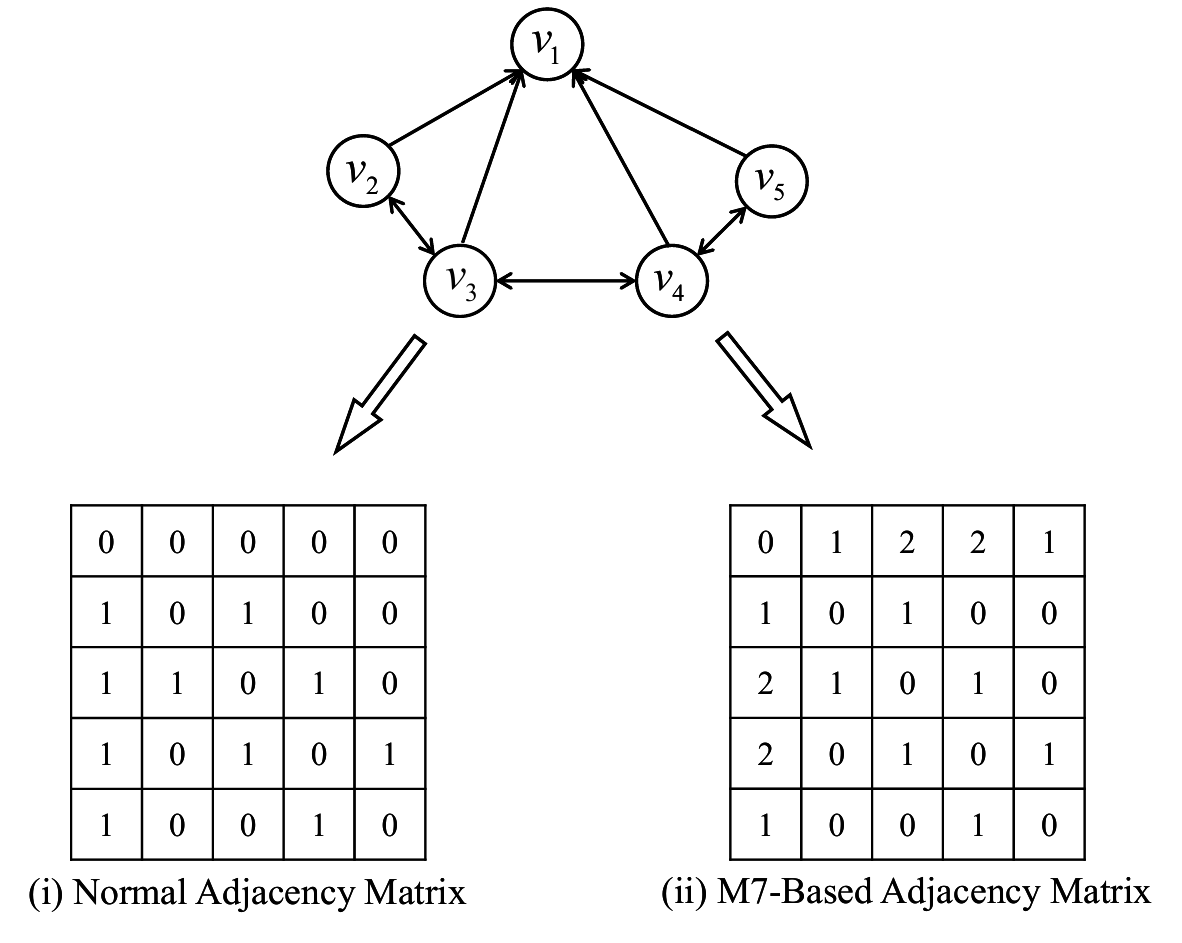}
        \caption{The Differences of Normal Adjacency Matrix and Motif-Based Adjacency Matrix.}
        \label{fig:sub2}
    \end{subfigure}
    \caption{Visualization of Seven Types of Motifs Considered and the Example for the Calculation of Motif-Based Adjacency Matrix.}
    \label{fig:main}
\end{figure*}

\subsection{Motif-Based Personalized PageRank} 
Given an unweighted graph $G$, the PageRank (PR) value represents the importance of each node, and the personalized PageRank (PPR) measures the importance range of each pair. From \autoref{eq1} and \autoref{eq2}, it is obtained that adjacency matrix $A$ can affect both $\textbf{P}$ and $\overset{\sim}{\mathop{\boldsymbol{A}}}\,'$, thus the final PR or PPR values. However, for both PR and PPR, the values are calculated based on the direct pair-wise relations, neglecting the higher-order relationships. To solve this problem, the concept of PageRank is combined with motif and yield motif-based personalized PageRank (MPPR). Motif can be regarded as a pattern of edges on a small number of nodes from a graph that can capture higher-order information in the network. PPR is a metric that can measure the importance of different nodes in the network. Thus, they can complement with each other, and the proposed MPPR is a metric that reflects the importance of different nodes in the network in consideration of the higher-order motifs. Note all of the motifs concerned in this paper refers to network motifs, rather than biology motifs. The following of this subsection introduces the calculation of MPPR. 

Following \cite{MotifDef, MPR}, given a motif set $\mathcal{M}\left(\boldsymbol{B}, \mathcal{A}\right)$, the motif-based adjacency matrix or co-occurrence matrix of a type of motif $M_i$ is defined as: 

\begin{equation}
A_{ij}^{M_i}=\sum\limits_{\left( v,{{\chi }_{_{\mathcal{A}}}}\left( v \right) \right)\in \mathcal{M}_i}{\mathbbm{1}\left( \left\{ i,j \right\}\subset {{\chi }_{_{\mathcal{A}}}}\left( v \right), i\neq j \right)}
\end{equation}
where $\mathbbm{1}\left(St \right) = 1$ if the boolean variable $St$ is $True$ and otherwise 0. The motif-based adjacency matrix $\boldsymbol{A}^{M_i}$ represents the frequency of two nodes appearing in a given motif. The higher $A_{ij}^{M_i}$ is, the more significant the relation between $i$ and $j$ is within the network in consideration of motif $M_i$. In this work, only three-node motifs are considered. And among the three-node motifs, we further restrict our focus on seven kinds of triangular motifs (\autoref{fig:sub1}) since the importance of them in networks has been verified by many studies \cite{triangular1, triangular2}. The computation results of motif-based adjacency matrices for $M_1$ to $M_7$ in \autoref{fig:main}(a) is grabbed from \cite{MotifDef, MPR}, which are presented in \autoref{appendix7motifs}. 

Take \autoref{fig:main}(b) as an example, the normal adjacency matrix of the network is given as (i), and motif-based adjacency matrix of $M_7$ motif is described as (ii). The corresponding values of $v_1\rightarrow v_3$ and $v_3\rightarrow v_1$ are both 2, because the edges $v_3\rightarrow v_1$ exists in two $M_7$-based motifs ($<v_1, v_3, v_4>$; $<v_1, v_2, v_3>$). The motif-based adjacency matrix is always symmetric. This is because it emphasizes how many motifs is an edge involved, rather than the dependence relationship of the linked nodes. In this paper, it is argued that the edge-based and motif-based relationships complement each other in measuring the relationships among the nodes. Thus, a simple yet effective linear transformation is applied to combine the edge-based and motif-based adjacency matrices: 

\begin{equation}\label{lineartransformation}
\boldsymbol{\vartheta}^{M_k} = \left( 1-\tau \right) \boldsymbol{A} + \tau \boldsymbol{A}^{M_k}
\end{equation}
where $\tau \in \left[0, 1\right]$ balances the edge-based and motif-based relations, $\boldsymbol{A}$ is the original adjacency matrix and $\boldsymbol{A}^{M_k}$ is the calculated motif-based adjacency matrix. By removing the $\overset{\sim}{\mathop{\boldsymbol{A}}}\,'$ in \autoref{eq2} by $\boldsymbol{\vartheta}^{M_k}$, the MPPR is calculated as, 
\begin{equation}\label{eq4}{{\Pi }_{\text{MPPR}}}=\alpha \cdot {{\left( {{\boldsymbol{I}}_{n}}-\left( 1-\alpha  \right)\boldsymbol{\vartheta}^{M_k} \right)}^{-1}} \end{equation}
which reflects the importance among nodes on the basis of considering higher-order motif relationships within the networks. 

\subsection{GCN-MPPR for Node Classification} 

Then, we consider a simple and widely used message passing algorithm for semi-supervised classification, graph convolutional network (GCN). In the case of two-layer message passing, the equation can be formulized as: 

\begin{equation}\label{eq5}{{Z}_{\text{GCN}}}=\text{softmax} \left( \overset{\sim}{\mathop{\boldsymbol{A}}}\,'\text{ReLU}\left( \overset{\sim}{\mathop{\boldsymbol{A}}}\,'X{{W}_{0}} \right){{W}_{1}} \right)\end{equation}
where $Z_{\text{GCN}} \in \mathbb{R}^{n\times c}$ is the predicted labels, $\overset{\sim}{\mathop{A}}\,'$ is discussed above and $W_1$ and $W_2$ are the trained weights. In such two-layer GCNs, only neighbors within the two-hop neighborhood are covered. This means that these neural networks fail to be so `deep'. In this work, MPPR is combined with MPNNs, which guides the propagation process at a `higher-order' level, resulting in a better and more promising propagation scheme. 

In order to utilize the preceding higher-order relationships for semi-supervised classification, predictions for each node are generated based on its own features and then they are propagated via MPPR introduced above to generate the final predictions. In such a way, the GCN-MPPR is founded, which results in changing \autoref{eq5} to: 
\begin{align}
  {{Z}_{\text{MPPR}}}&= \left( \alpha \cdot {{\left( {{\boldsymbol{I}}_{n}}-\left( 1-\alpha  \right)\boldsymbol{\vartheta}_{M_k} \right)}^{-1}} \right)^\beta H  \nonumber \\ 
   H&= f_\theta\left(X\right), \quad H \in \mathbb{R}^{|V| \times C}.
\end{align}
where $\boldsymbol{X}$ is the feature matrix and $f_{\theta}$ is a neural network with parameter set $\theta$, $\boldsymbol{H} \in \mathbb{R}^{n\times c}$ is the generated predictions and $\beta$ is defined as a slipping factor.   

The final class probabilities are obtained by applying a row-wise softmax,
\begin{equation}
P = \mathrm{softmax}(Z).
\end{equation}
Only a small subset of nodes $\mathcal{V}_{\mathrm{train}} \subset V$ is labeled during training. The model is trained end-to-end by minimizing the cross-entropy loss over labeled nodes:
\begin{equation}
\mathcal{L}_{\mathrm{node}} =
- \sum_{v \in \mathcal{V}_{\mathrm{train}}}
\sum_{c=1}^{C}
Y_{vc} \log P_{vc}.
\end{equation}
Gradients are backpropagated through both the neural network $f_{\theta}$ and the propagation operator, enabling the model to implicitly leverage information from large graph neighborhoods while avoiding oversmoothing.

Similar to PPNP \cite{ppnp}, GCN-MPPR separates the neural network used for generating predictions from the propagation scheme. This separation thus solves the second problem discussed in \autoref{intro}. The depth of the neural network is now fully independent of the propagation algorithm. It is important to note that GCN-MPPR is also trained end-to-end. That is, the gradient flows through the propagation scheme during backpropagation (implicitly considering infinitely many neighborhood aggregation layers). Compared to PPNP, however, the proposed GCN-MPPR solves the propagation issue by motif-based personalized PageRank, which considers the higher-order relationships of nodes in the considered network. Adding the propagation effects based on MPPR significantly upgrades the model’s performance.

\subsection{GCN-MPPR for Link Prediction}

Then, we introduce the GCN-MPPR framework for link prediction.  We stick to the upmentioned notations and definations to study link prediction based on GCN-MPPR. $X \in \mathbb{R}^{|V|\times d}$. To avoid information leakage, we first split observed edges into training/validation/test sets. All message passing and propagation are performed on the training graph $G_{\mathrm{train}}=(V,E_{\mathrm{train}})$, which is obtained by removing the validation and test positive edges from $E$. For supervision, we pair each mini-batch of positive edges with an equal number of negative samples (non-edges) and optimize a binary classification objective.

Our encoder follows the Personalized PageRank Neural Network (PPNP) principle by decoupling feature transformation from graph propagation. Specifically, we compute local node representations with a two-layer multilayer perceptron (MLP) $f_{\theta}$ with ReLU activations and dropout:
\begin{equation}
H = f_{\theta}(X).
\end{equation}

We then diffuse these representations over $G_{\mathrm{train}}$ using a motif-based personalized pageRank operator with teleport (restart) probability $\alpha \in (0,1)$.
Let $\tilde{A}$ denote the (symmetrically) normalized adjacency matrix of $G_{\mathrm{train}}$. The PPR propagation can be written in closed form as
\begin{equation}
Z = {\Pi }_{\text{MPPR}} H
= \left( \alpha \cdot {{\left( {{\boldsymbol{I}}_{n}}-\left( 1-\alpha  \right)\boldsymbol{\vartheta}^{M_k} \right)}^{-1}} \right) ^\beta H,
\end{equation}

Given the final node embeddings $Z$, we score a candidate edge $(u,v)$ using a dot-product decoder, which is then mapped into an existence probability with the activation function (logistic sigmoid as an example in this paper): 
\begin{equation}
p_{uv} = \sigma(z_u^{\top} z_v) = \frac{1}{1+\exp(-s_{uv})}.
\end{equation}

The model is trained end-to-end with binary cross-entropy on positive edges and negative samples:
\begin{equation}
\mathcal{L} =
-\sum_{(u,v)\in \mathcal{E}^{+}} \log p_{uv}
-\sum_{(u,v)\in \mathcal{E}^{-}} \log (1-p_{uv}),
\end{equation}
where $\mathcal{E}^{+}$ and $\mathcal{E}^{-}$ denote the sets of positive and negative training pairs, respectively. 

\section{Experimental Results and Analysis}\label{sec4}
In this section, comparative evaluations of GCN-MPPR against baselines are first performed on well-designed semi-supervised node classification and link prediction tasks. Then, GCN-MPPR is applied in a latest GCN-related framework named DGCRL to estimate how much GCN-MPPR would benefit for it. 

\begin{table*}[!t]
\renewcommand{\arraystretch}{0.98}
\caption{The Accuracy Results of Node Classification via GCN-MPPR and its competitors. \label{table-acc}}
\vspace{-8pt}
\setlength{\tabcolsep}{5mm}
\begin{tabular}{lcccc}
\hline
           & Cora & PubMed & Amazon\_computers & Amazon\_photo \\ \hline
GCN        & 0.7836${\scriptstyle \pm 0.0195}$ & 0.7765${\scriptstyle \pm 0.4000}$ & 0.7422${\scriptstyle \pm 0.2350}$ & 0.9023${\scriptstyle \pm 0.1088}$ \\
GAT        & 0.7908${\scriptstyle \pm 0.0105}$ & 0.7776${\scriptstyle \pm 0.4400}$ & 0.7662${\scriptstyle \pm 0.5615}$ & 0.8942${\scriptstyle \pm 0.2062}$ \\
GIN        & 0.7605${\scriptstyle \pm 0.0805}$ & 0.7695${\scriptstyle \pm 0.3301}$ & 0.7588${\scriptstyle \pm 0.8142}$ & 0.8901${\scriptstyle \pm 0.4468}$ \\
PPNP       & 0.8247${\scriptstyle \pm 0.0118}$ & 0.8021${\scriptstyle \pm 0.0172}$ & 0.7722${\scriptstyle \pm 0.1400}$ & 0.9113${\scriptstyle \pm 0.0649}$ \\
APPNP      & 0.8234${\scriptstyle \pm 0.0121}$   & 0.8009${\scriptstyle \pm 0.0173}$  &  0.7664${\scriptstyle \pm 0.1518}$ & 0.9121${\scriptstyle \pm 0.0618}$    \\
PPRGo      & 0.8201${\scriptstyle \pm 0.2563}$ & 0.7148${\scriptstyle \pm 0.1151}$ & 0.6528${\scriptstyle \pm 0.5540}$ & 0.6990${\scriptstyle \pm 0.6532}$ \\
SHP-GNN    & 0.7721${\scriptstyle \pm 0.0325}$ & \textbf{0.8058${\scriptstyle \pm 0.0725}$}  & 0.7656${\scriptstyle \pm 0.0725}$  & 0.9008${\scriptstyle \pm 0.0332}$  \\ \hline
GCN-MPPR$_{M1}$       & \textbf{0.8276${\scriptstyle \pm 0.0123}$}  & 0.7935${\scriptstyle \pm 0.0138}$  & \textbf{0.8467${\scriptstyle \pm 0.0137}$} & \textbf{0.9345${\scriptstyle \pm 0.0061}$} \\ 
GCN-MPPR$_{M2}$      & \textbf{0.8253${\scriptstyle \pm 0.0131}$} & 0.7940${\scriptstyle \pm 0.0126}$ &  \textbf{0.8468${\scriptstyle \pm 0.0139}$} & \textbf{0.9336${\scriptstyle \pm 0.0066}$} \\ 
GCN-MPPR$_{M3}$      & \textbf{0.8265${\scriptstyle \pm 0.0096}$}  & 0.7941${\scriptstyle \pm 0.0125}$ & \textbf{0.8479${\scriptstyle \pm 0.0126}$} & \textbf{0.9324${\scriptstyle \pm 0.0058}$} \\ 
GCN-MPPR$_{M4}$      & 0.8163${\scriptstyle \pm 0.0106}$ & 0.7938${\scriptstyle \pm 0.0176}$ & \textbf{0.8475${\scriptstyle \pm 0.0126}$} & \textbf{0.9295${\scriptstyle \pm 0.0056}$} \\ 
GCN-MPPR$_{M5}$       & \textbf{0.8273${\scriptstyle \pm 0.0109}$} & 0.7938${\scriptstyle \pm 0.0124}$ & \textbf{0.8479${\scriptstyle \pm 0.0132}$} & \textbf{0.9333${\scriptstyle \pm 0.0059}$} \\ 
GCN-MPPR$_{M6}$       & \textbf{0.8302${\scriptstyle \pm 0.0106}$} & 0.7937${\scriptstyle \pm 0.0130}$ & \textbf{0.8468${\scriptstyle \pm 0.0130}$} & \textbf{0.9339${\scriptstyle \pm 0.0066}$} \\ 
GCN-MPPR$_{M7}$       & \textbf{0.8334${\scriptstyle \pm 0.0089}$} & 0.7942${\scriptstyle \pm 0.0125}$ & \textbf{0.8453${\scriptstyle \pm 0.0140}$} & \textbf{0.9316${\scriptstyle \pm 0.0066}$} \\
GCN-MPPR$_{Mean}$     & \textbf{0.8266}   & 0.7939  & \textbf{0.8470}  & \textbf{0.9326} \\ 
\hline
\end{tabular}
\end{table*}

\subsection{Datasets and Baselines} Four datasets are applied for node classification and link prediction tasks, namely Cora \cite{cora}, PubMed \cite{pubmed}, Amazon computers and Amazon photo \cite{amazon}. The details of these datasets are presented in \autoref{appendixdataset}. Basic models including GCN \cite{gcn} and GAT \cite{gat} are selected as baselines for both tasks. For node classification, the baselines also contain GIN \cite{GIN}, PPNP (\& APPNP) \cite{ppnp}, PPRGo \cite{PPRGo}, SHP-GNN \cite{shp-GNN}. For link prediction, GAE (\& VGAE) \cite{VGAE} and HDGL \cite{HDGL} are considered for evaluation. 

\subsection{Node Classification Results} In order to verify the advantages of GCN-MPPR, we first utilize a well-designed semi-supervised node classification task to test the performance of it with its competitors. 

\subsubsection{Experimental Settings} In the experiments, each dataset is divided into a visible training set and an invisible test set, which do not change. The test set was only used once to report the final performance.  Additionally, the $early$ $stopping$ criterion the same as PPNP and APPNP is applied across models, whose details can be referred to \cite{ppnp}. To maintain fairness, the neural network for GCN-MPPR is structurally very similar to GCN and PPNP and has the same number of parameters. Two layers with $h$ = 64 hidden units are used. To avoid overfitting, we apply $L_2$-regularization with $\lambda = 0.005$ on the weights of the first layer and use dropout with dropout rate $d = 0.5$ on both layers and the adjacency matrix. For the teleport probability of MPPR, $\alpha = 0.1$. The slipping factor $\beta$ for $Cora$ and $Pubmed$ is set as 0.5, and 0.75 for the other datasets. The parameter $\tau$ applied in \autoref{lineartransformation} is set as 0.9. Each experiment is conducted 100 times on multiple random splits. The average and variance of accuracy are reported.  

\subsubsection{Classification Accuracy Results} The classification accuracy of various models is presented in \autoref{table-acc}. The results of GCN-MPPR are presented in bold if it outperforms all of the baselines.  First, it is obvious that almost all of the classification accuracy of GCN-MPPR is higher than all of the baselines, no matter which type of motif structure is considered (from $M_1$ to $M_7$). Second, the performance of GCN-MPPR on PubMed network is not so competitive. This could be partly attributed to the fact that the motif within this network is less meaningful, which even poses a negative effect of the classification accuracy on GCN-MPPR. However, even in this case, the gap between GCN-MPPR and the best performer on PubMed network is quite narrow. Meanwhile, the training process of GCN-MPPR saves more than 60\% time than many the baselines (further discussed in \autoref{subsubtime}). Furthermore, the variance of GCN-MPPR is significantly lower than the competitors. This means that the addition of MPPR on GCN improves the stability of traditional GCN models. As a consequence, there are enough evidences to believe in the performance of GCN-MPPR. The box plots of 100-run experiments on various datasets are presented in \autoref{appendixbox}, which further demonstrate the stability of GCN-MPPR.  

\subsubsection{Classification Runtime Results} \label{subsubtime} The runtime analysis is conducted on Amazon Computers network, and results are shown in \autoref{runtime}. The best performer on runtime is shown in bold, and the second-best performer is presented underlined.  It could be seen that GCN-MPPR is much faster than all of the baselines except PPRGo on the total time and the training time per epoch. This is because, with the addition of MPPR, it can accelerate the propagation process greatly by propagating the information deeper and more proper. In this way, the $early$ $stopping$ condition is achieved faster. It is important to emphasize that the approximate PPNP saves a lot of time during the calculation of PPR than PPNP, rather than saves the time during training. Thus, it is reasonable to find that its training time is even longer than PPNP, since \autoref{runtime} considers the runtime comparison only. Though PPRGo is fast than GCN-MPPR, its performance is yet limited as is presented in \autoref{table-acc} and mentioned above. This shows that PPRGo is extremely competitive when scaling MPNNs to networks with millions of nodes, rather than making accurate classifications, which agrees with the birth of PPRGo. As for the training time per epoch, GCN-MPPR, PPNP and APPNP are about slower than GCN, which is mainly because of the larger number of matrix multiplications. Meanwhile, the training time per epoch of GCN-MPPR is less than the ones of PPNP and APPNP. This could be partially attributed to the fact that the calculated MPPR is sparser than PPR, thereby reducing the time of matrix multiplications. It worth emphasizing that GCN-MPPR is faster than other models such as PPNP mainly because it can lead the training process wiser and more proper due to the higher-order motif consideration, such that the $early$ $stopping$ condition is achieved faster. 

\begin{table}[!t]
\renewcommand{\arraystretch}{0.98}
\caption{Time Consumption Analysis During Propagation on Amazon Computers Network. \label{runtime}}
\vspace{-8pt}
\begin{tabular}{lccc}
\hline
           & Total Time & Last Epoch & Average Per Epoch    \\ \hline
GCN        & 102.18s & 3050 & 33.50ms   \\
GAT        & 178.81s & 1980 & 90.26ms  \\
PPNP       & 75.85s & 1715 &  44.22ms  \\
APPNP      & 96.91s & 2069 &  46.88ms  \\
PPRGo      & \textbf{6.59s} & 200 & \textbf{32.95ms}   \\
SHC-Concat & 59.71s & 820 & 72.80ms   \\ \hline
GCN-MPPR       & \ul{ 29.10s} & 860 &  \ul{ 33.83ms}  \\ 
\hline
\end{tabular}
\end{table}

\subsubsection{Hyper-Parameter Analysis} 
\begin{figure}[!t]
    \centering
    \begin{subfigure}[t]{1\columnwidth}
        \centering
        \includegraphics[width=1.02\textwidth]{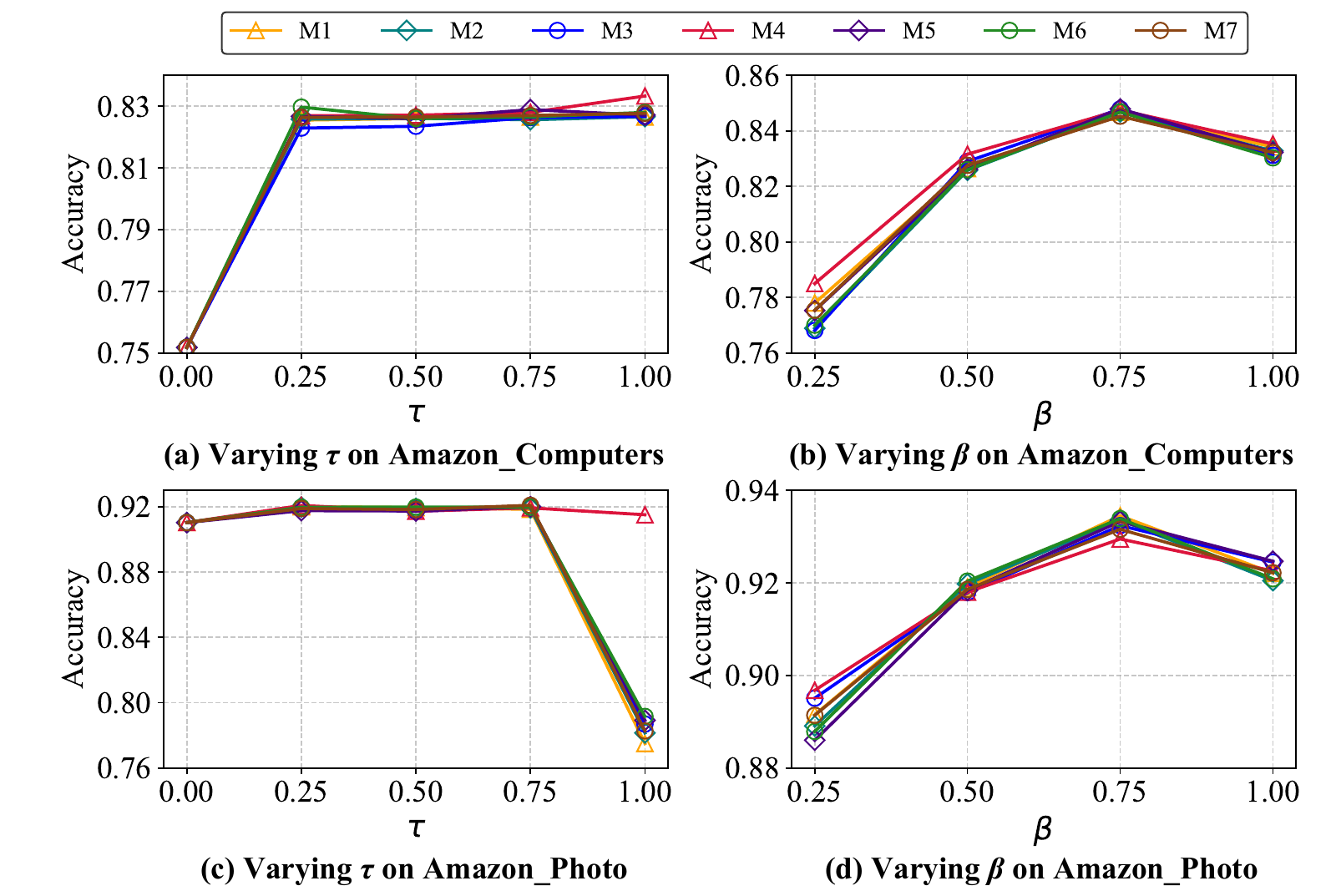}
    \end{subfigure}
    \vspace{-16pt}
    \caption{Accuracy when varying parameters.}
    \label{fig-hp}
\end{figure}

For the model, this work has three hyper-parameters, $\tau$, $\alpha$, and $\beta$. Since the reasonable selection of $\alpha$ has already been fully discussed in related references for PageRank \cite{PageRank, Zebra, PR_add}, we do not discuss it in this article. We conduct experiments regarding the setting of $\tau$ and $\beta$ respectively to test the parameter sensitivity of the model on $Amazon\_computers$ dataset. First, we set $\tau \in [0, 0.25, 0.5, 0.75, 1]$. Noticeably, when $\tau = 0$, the GCN-MPPR degenerates into PPNP, and when $\tau = 0$ the linear relationships in \autoref{lineartransformation} is totally neglected. As shown in the \autoref{fig-hp}(a) and \ref{fig-hp}(c), compared to $\tau = 0$ or $\tau = 1$, the performance of GCN-MPPR is significantly better, which demonstrates that the complementary effect of \autoref{lineartransformation} and utilization of MPPR is effective. Meanwhile, it shows that GCN-MPPR is almost robust throughout different settings of $\tau$ except $\tau = 0$ or $\tau = 1$. Then, as is shown in \autoref{fig-hp}(b) and \ref{fig-hp}(d), when $\beta$ rises from 0.25 to 0.75, the performance of GCN-MPPR increases, though the increase rate drops a little after $\beta$ reaches 0.5. After that, the performance of GCN-MPPR drops a little as $\beta$ increases from 0.75 to 1. This is mainly because that the calculated MPPR matrix is relatively sparse. Thus, we need $\beta$-power $\left( \beta < 0 \right)$ to enlarge and activate some elements in the matrix. However, the setting of $\beta$ should not be too little yet. According to the hyper-parameter analysis, we observe the ideal setting of $\beta$ should range from 0.5 to 0.75. 

\subsubsection{Ablation Study} The ablation study is carried out in both ways on $Amazon$ $Computers$ Network. One the one hand, we remove the motif and MPPR on GCN-MPPR respectively. One the other hand, as the MPPR propagation scheme affecting (a) the neural network during training, and (b) the classification decision during predicting, it is worthwhile to investigate how the model performs without propagation during (a), (b) or both of them. \autoref{ablation} shows whether MPPR is effective and whether the MPPR-based propagation affects both training and predicting. First, he optimal results come from the full propagation of GCN-MPPR, which validates our approach. And the performance of propagation during prediction only is quite satisfactory for both PPNP and GCN-MPPR. This re-verifies the conclusion in \cite{ppnp} that these models can be combined with pre-trained models that do not incorporate any graph information and still significantly improve their accuracy. It also shows that just propagating during training can also result in improvements. This indicates that GCN-MPPR can also be applied to online/inductive learning where only the features and not the neighborhood information of an incoming (previously unobserved) node is available.

\begin{table}[!t]\centering
\caption{The Results of Ablation Study.\label{ablation}}
\vspace{-8pt}
\renewcommand{\arraystretch}{0.98}
\begin{tabular}{l|cccc}\hline
                 & None & Train & Predict & Train \& Predict \\\hline
w/o MPPR         & \multicolumn{4}{c}{69.0\%}                 \\\hline
w/o Motif (PPNP) & 69.0\%     & 71.5\%  &  75.8\% &    77.2\%        \\
GCN-MPPR$_{Mean}$ & 69.0\%     &71.5\%       &  76.3\%  &    82.6\%         \\ \hline
\end{tabular}
\end{table}

\subsection{Link Prediction Results} 

\subsubsection{Experimental Settings} 
In case of link prediction experiments, all datasets are firstly preprocessed by selecting the largest connected component and removing self-loops. Node features are used in their original bag-of-words or attribute form without additional normalization or dimensionality reduction. Positive edge samples are directly taken from the undirected edges present in each graph. We adopt a transductive random edge splitting protocol. For each independent run, train/validation/test = 5:1:4; To ensure the training graph remains connected, we first extract a BFS spanning tree and force all its edges into the training set. The remaining edges are then randomly allocated to validation, test, or training sets. Negative samples are generated at a 1:1 ratio (one negative per positive edge). All negatives are sampled from the full original graph to avoid including any true existing edges. For GCN, PPNP and GCN-MPPR, we adopt classic 2-layer GCNs, using a hidden dimension of 64 and an output embedding dimension of 64. Training is performed with the Adam optimizer (learning rate 1e-3 in all experiments). The maximum number of training epochs ranges from 1000 to 25000 depending on dataset size, without early stopping setting. Batch size is set to 1024. All experiments are repeated 100 times, and the mean AUC and AP (Average Precision) are reported. 
\subsubsection{Link Prediction Results} 

As is shown in \autoref{table-LP}, no matter which kind of motif is considered, the AUC and AP of GCN-MPPR is very competitive. The results further demonstrate that GCN-MPPR could propagate the information deeper and further without adding the number of layer, benefitting from the utlization of MPPR. Noticeably, the AUC and AP on $Cora$ dataset surpass all baselines. This means that GCN-MPPR is still competitive when training set is limited on small-scale networks. On both $PubMed$ and $Amazon\_Photo$ datasets, GCN-MPPR outperforms the baselines on AUC, while performs worse then them on AP. This could be mainly attributed to the fact that MPPR tends to emphasize the global higher-order information capture, which lacks the fine-grained ranking refinement, thus hurting AP. However, even in such case the gap between GCN-MPPR and its competitors is quite narrow. 

\small
\begin{table}[!t]
\renewcommand{\arraystretch}{0.98}
\caption{The Results of Link Prediction via GCN-MPPR and its competitors. \label{table-LP}}
\vspace{-8pt}
\begin{tabular}{lcccccc}
\hline
                   & \multicolumn{2}{c}{Cora} & \multicolumn{2}{c}{PubMed} & \multicolumn{2}{c}{Amazon\_Photo} \\ \cline{2-7} 
                   & AUC         & AP         & AUC          & AP          & AUC              & AP             \\ \hline
GCN                & 0.8822      &  0.8578    & 0.9582       & 0.9314      &  0.9799          &  0.9394        \\
GAT         &     0.8722         &   0.8362  &  0.9356           &0.9214       &    0.9725      &  0.9456                  \\
PPNP               & 0.8918      & 0.8657     & 0.9606       & 0.9317      &  0.9777          &  0.9351        \\
GAE                & 0.8184      & 0.8514     & 0.9535       & \textbf{0.9597}      &  0.9745          &  \textbf{0.9740}        \\
VGAE               & 0.7490      & 0.7770     & 0.9469       & 0.9533      &  0.9696          &  0.9690        \\
HDGL               &   0.6720      &      0.6862  &   0.8513 & 0.8838      &        0.7617   &  0.7706       \\ \hline
\small{GCN-MPPR$_{M1}$}   &  \textbf{0.8965} & \textbf{0.8752}  & \textbf{0.9653} & 0.9375  & \textbf{0.9800}     &   0.9389       \\
\small{GCN-MPPR$_{M2}$}   &  \textbf{0.8973} & \textbf{0.8760}  & \textbf{0.9653} & 0.9376  & \textbf{0.9802}     &   0.9390       \\
\small{GCN-MPPR$_{M3}$}   &  \textbf{0.8967} & \textbf{0.8751}  & \textbf{0.9654} & 0.9376  & 0.9798     &   0.9387       \\
\small{GCN-MPPR$_{M4}$}   &  \textbf{0.8921} & \textbf{0.8695}  & \textbf{0.9643} & 0.9356  & 0.9785     &   0.9383       \\
\small{GCN-MPPR$_{M5}$}   &  \textbf{0.8968} & \textbf{0.8755}  & \textbf{0.9654} & 0.9377  & \textbf{0.9799}     &   0.9387       \\
\small{GCN-MPPR$_{M6}$}   &  \textbf{0.8974} & \textbf{0.8761}  & \textbf{0.9652} & 0.9374  & \textbf{0.9800}     &   0.9389        \\
\small{GCN-MPPR$_{M7}$}   &  \textbf{0.8977} & \textbf{0.8757}  & \textbf{0.9652} & 0.9374  & \textbf{0.9800}     &   0.9388      \\ \hline
\small{GCN-MPPR$_{Mean}$} &  \textbf{0.8964} & \textbf{0.8747}  & \textbf{0.9652} & 0.9373  & \textbf{0.9799}     &   0.9388      \\ \hline
\end{tabular}
\end{table}
\normalsize

\subsection{GCN-MPPR into other GCN Framework} 

GCN-MPPR shows the potential to be generalized in typical GCN tasks. As long as a part of a complicated task involves the use of any traditional GCN model to obtain representations, then by introducing the propagation optimization based on MPPR, theoretically, the performance of this GCN can be improved, thereby enhancing the overall performance of the task. To verify this, we conducted experiments on the following recently proposed framework that incorporates the GCN module. We replaced the GCN module in the overall framework with our GCN-MPPR module and verified whether the performance had improved due to the introduction of MPPR. 

Consider the recent proposed Directed Graph Contrastive Representation Learning (DGCRL) framework for link prediction (especially for the reconstruction for gene regulatory networks, GRNs in short). The details of DGCRL can be referred to \cite{DGCRL} and the details of the GCN-MPPR implementation on it can be referred to \autoref{appendixDGCRL}. Simply speaking, the simple GCN in DGCRL is replaced by GCN-MPPR to learn the representation better. The GRN reconstruction results are as shown in \autoref{tableothers} (w. means with). No matter which type of motif is considered, the overall reconstruction performance increases a lot, with both AUC and AP (Average Precision) upgrading above 7\% on in silico and 2\% on S. cerevisiae dataset, respectively.  Furthermore, it should be noted that, in such phenomenon, if the early stopping setting is removed and the training epoch is set to be extremely large, the performance of DGCRL with GCN-MPPR degrades significantly, and the decline is much greater than that of DGCRL. This might suggest that GCN-MPPR has some shortcomings in dealing with overfitting. The cause of this issue is still under research and might be explained by future research. However, our results are still sufficient to confirm the outstanding performance and application prospects of GCN-MPPR in many GCN tasks. 

\begin{table}[!t]
\caption{GCN-MPPR into DRCRL. \label{tableothers}}
\vspace{-8pt}
\centering
\renewcommand{\arraystretch}{0.98}
\begin{tabular}{ccccc}
\hline
              & \multicolumn{2}{c}{in silico} & \multicolumn{2}{c}{S. cerevisiae} \\ \cline{2-5} 
              & AUC    & AP  & AUC       & AP       \\ \hline
DGCRL         &     0.7813         &   0.7548        & 0.8194        &  0.7857           \\
DGCRL w. MPPR$_{M1}$ & \textbf{0.8438}      &   \textbf{0.8264}        & \textbf{0.8410}        &  \textbf{0.7986}           \\
DGCRL w. MPPR$_{M2}$ & \textbf{0.8397}      &   \textbf{0.8207}        & \textbf{0.8415}        &  \textbf{0.8021}           \\
DGCRL w. MPPR$_{M3}$ & \textbf{0.8418}      &   \textbf{0.8229}        & \textbf{0.8390}        &  \textbf{0.8000}           \\
DGCRL w. MPPR$_{M4}$ & \textbf{0.8426}      &   \textbf{0.8214}        & \textbf{0.8359}        &  \textbf{0.7915}           \\
DGCRL w. MPPR$_{M5}$ & \textbf{0.8412}      &   \textbf{0.8223}        & \textbf{0.8394}        &  \textbf{0.7969}           \\
DGCRL w. MPPR$_{M6}$ & \textbf{0.8425}      &   \textbf{0.8267}        & \textbf{0.8403}        &  \textbf{0.7989}           \\
DGCRL w. MPPR$_{M7}$ & \textbf{0.8430}      &   \textbf{0.8264}        & \textbf{0.8407}        &  \textbf{0.7961}           \\ \hline
{\it Improvement} & $\geq 7.47\%$ & $\geq8.73\% $ & $\geq2.01\% $ &$\geq1.64\% $ \\\hline
\end{tabular}
\end{table}

\section{Conclusion}\label{sec5}
Graph is a ubiquitous structure that shows promise for many applications \cite{add00001, add00002}. MPNNs and their variants have made great successes over the past decades. However, the propagation performance of message passing neural networks are usually limited due to the fact that the propagation depth is not `deep enough', leading to a series of works to solve the problem. Most existing methods tend to solve the problems via linear metrics, which are yet limited to capture the higher-order structure relationships within the networks. As a result, most existing GCN models still suffer from low accuracy, limited stability and high computation cost. To overcome these challenges, MPPR is proposed as a novel variant of PageRank to record the influence of one node to another on the basis of considering higher-order motif relationships. Secondly, the MPPR is utilized to the message passing process of GCNs, thereby avoiding over-smoothing during the message passing process. The experimental results show that the proposed method outperforms almost all of the baselines on accuracy, stability and time consumption. Additionally, the proposed method can be considered as a component, that can be applied to almost all GCN tasks, with DGCRL being demonstrated in the experiments. 



\newpage

\bibliographystyle{unsrt}
\bibliography{sample-base}

@String{Computing = "Computing" }

@String{Computer = "{IEEE} Computer" }

@String{Springer = "Springer-Verlag" }

@article{motif2,
title = {Motif-based Contrastive Graph Clustering with clustering-oriented prompt},
journal = {Information Processing \& Management},
volume = {62},
number = {5},
pages = {104208},
year = {2025},
issn = {0306-4573},
doi = {https://doi.org/10.1016/j.ipm.2025.104208},
url = {https://www.sciencedirect.com/science/article/pii/S0306457325001499},
author = {Xunlian Wu and Jingqi Hu and Yining Quan and Qiguang Miao and Peng Gang Sun},
}

@inproceedings{PPRGo,
  author       = {Aleksandar Bojchevski and
                  Johannes Klicpera and
                  Bryan Perozzi and
                  Amol Kapoor and
                  Martin Blais and
                  Benedek R{\'{o}}zemberczki and
                  Michal Lukasik and
                  Stephan G{\"{u}}nnemann},
  editor       = {Rajesh Gupta and
                  Yan Liu and
                  Jiliang Tang and
                  B. Aditya Prakash},
  title        = {Scaling Graph Neural Networks with Approximate PageRank},
  booktitle    = {{KDD} '20: The 26th {ACM} {SIGKDD} Conference on Knowledge Discovery
                  and Data Mining, Virtual Event, CA, USA, August 23-27, 2020},
  pages        = {2464--2473},
  publisher    = {{ACM}},
  year         = {2020},
  url          = {https://doi.org/10.1145/3394486.3403296},
  doi          = {10.1145/3394486.3403296},
}

@inproceedings{ppnp,
  author       = {Johannes Klicpera and
                  Aleksandar Bojchevski and
                  Stephan G{\"{u}}nnemann},
  title        = {Predict then Propagate: Graph Neural Networks meet Personalized PageRank},
  booktitle    = {7th International Conference on Learning Representations, {ICLR} 2019,
                  New Orleans, LA, USA, May 6-9, 2019},
  publisher    = {OpenReview.net},
  year         = {2019},
  url          = {https://openreview.net/forum?id=H1gL-2A9Ym},
}

@article{PPRTGI,
  author       = {Ke Ma and
                  Jiawei Li and
                  Mengyuan Zhao and
                  Ibrahim Zamit and
                  Bin Lin and
                  Fei Guo and
                  Jijun Tang},
  title        = {{PPRTGI:} {A} Personalized PageRank Graph Neural Network for TF-Target
                  Gene Interaction Detection},
  journal      = {{IEEE} {ACM} Trans. Comput. Biol. Bioinform.},
  volume       = {21},
  number       = {3},
  pages        = {480--491},
  year         = {2024},
  url          = {https://doi.org/10.1109/TCBB.2024.3374430},
  doi          = {10.1109/TCBB.2024.3374430},
}

@article{Zebra,
  author       = {Yiming Li and
                  Yanyan Shen and
                  Lei Chen and
                  Mingxuan Yuan},
  title        = {Zebra: When Temporal Graph Neural Networks Meet Temporal Personalized
                  PageRank},
  journal      = {Proc. {VLDB} Endow.},
  volume       = {16},
  number       = {6},
  pages        = {1332--1345},
  year         = {2023},
  url          = {https://www.vldb.org/pvldb/vol16/p1332-li.pdf},
  doi          = {10.14778/3583140.3583150},
}

@article{DGCRL,
title = {Inferring Gene Regulatory Networks via Directed Graph Contrastive Representation Learning},
journal = {Knowledge-Based Systems},
volume = {316},
pages = {113324},
year = {2025},
issn = {0950-7051},
doi = {https://doi.org/10.1016/j.knosys.2025.113324},
url = {https://www.sciencedirect.com/science/article/pii/S0950705125003715},
author = {Kaifu Long and Luxuan Qu and Weiyiqi Wang and Zhiqiong Wang and Mingcan Wang and Junchang Xin},
}

@inproceedings{RS3,
  author       = {Jiahui Hu and
                  Jie Xu and
                  Jiakun Chen and
                  LiQiang Qiao and
                  Jilu Wang and
                  Feiran Huang and
                  Chaozhuo Li},
  editor       = {Rafik Hadfi and
                  Patricia Anthony and
                  Alok Sharma and
                  Takayuki Ito and
                  Quan Bai},
  title        = {Detaching Range from Depth: Personalized Recommendation Meets Personalized
                  PageRank},
  booktitle    = {{PRICAI} 2024: Trends in Artificial Intelligence - 21st Pacific Rim
                  International Conference on Artificial Intelligence, {PRICAI} 2024,
                  Kyoto, Japan, November 18-24, 2024, Proceedings, Part {I}},
  series       = {Lecture Notes in Computer Science},
  volume       = {15281},
  pages        = {454--466},
  publisher    = {Springer},
  year         = {2024},
}

@article{RS1,
  author       = {Cataldo Musto and
                  Pasquale Lops and
                  Marco de Gemmis and
                  Giovanni Semeraro},
  title        = {Context-aware graph-based recommendations exploiting Personalized
                  PageRank},
  journal      = {Knowl. Based Syst.},
  volume       = {216},
  pages        = {106806},
  year         = {2021},
  url          = {https://doi.org/10.1016/j.knosys.2021.106806},
  doi          = {10.1016/J.KNOSYS.2021.106806}
}

@article{MotifDef,
author = {Austin R. Benson  and David F. Gleich  and Jure Leskovec },
title = {Higher-order organization of complex networks},
journal = {Science},
volume = {353},
number = {6295},
pages = {163-166},
year = {2016},
doi = {10.1126/science.aad9029},
URL = {https://www.science.org/doi/abs/10.1126/science.aad9029},
eprint = {https://www.science.org/doi/pdf/10.1126/science.aad9029}}

@article{ SGCNN1,
  author       = {Yang Yi and
                  Xuequan Lu and
                  Shang Gao and
                  Antonio Robles{-}Kelly and
                  Yuejie Zhang},
  title        = {Graph classification via discriminative edge feature learning},
  journal      = {Pattern Recognit.},
  volume       = {143},
  pages        = {109799},
  year         = {2023}
}

@inproceedings{ SGCNN2,
  author       = {Jingwei Guo and
                  Kaizhu Huang and
                  Xinping Yi and
                  Rui Zhang},
  title        = {Graph Neural Networks with Diverse Spectral Filtering},
  booktitle    = {{WWW}},
  pages        = {306--316},
  publisher    = {{ACM}},
  year         = {2023}
}

@inproceedings{ SGCNN3,
  author       = {Mingguo He and
                  Zhewei Wei and
                  Shikun Feng and
                  Zhengjie Huang and
                  Weibin Li and
                  Yu Sun and
                  Dianhai Yu},
  title        = {Spectral Heterogeneous Graph Convolutions via Positive Noncommutative
                  Polynomials},
  booktitle    = {{WWW}},
  pages        = {685--696},
  publisher    = {{ACM}},
  year         = {2024}
}

@article{ MP1,
  author       = {Tiantian He and
                  Yang Liu and
                  Yew{-}Soon Ong and
                  Xiaohu Wu and
                  Xin Luo},
  title        = {Polarized message-passing in graph neural networks},
  journal      = {Artif. Intell.},
  volume       = {331},
  pages        = {104129},
  year         = {2024}
}

@article{ MP2,
  title={Aggregation or separation? Adaptive embedding message passing for knowledge graph completion},
  author={Li, Zhifei and Chen, Lifan and Jian, Yue and Wang, Han and Zhao, Yue and Zhang, Miao and Xiao, Kui and Zhang, Yan and Deng, Honglian and Hou, Xiaoju},
  journal={Information Sciences},
  volume={691},
  pages={121639},
  year={2025},
  publisher={Elsevier}
}

@inproceedings{ MP3,
  title={Revisiting neighborhood aggregation in graph neural networks for node classification using statistical signal processing},
  author={Ghogho, Mounir},
  booktitle={ICASSP 2025-2025 IEEE International Conference on Acoustics, Speech and Signal Processing (ICASSP)},
  pages={1--5},
  year={2025},
  organization={IEEE}
}

@article{NE01,
  author       = {Yong Tang and
                  Zhongming Xie and
                  Yuhan Fan and
                  Kaiyang Zhong and
                  Muhammet Deveci},
  title        = {Inferring anti-viral drugs using nonnegative matrix factorization
                  with self-paced learning and hypergraph regularization},
  journal      = {Expert Syst. Appl.},
  volume       = {296},
  pages        = {129104},
  year         = {2026},
  url          = {https://doi.org/10.1016/j.eswa.2025.129104},
  doi          = {10.1016/J.ESWA.2025.129104},
}

@article{NE02,
  author       = {Wang Zhou and
                  Zhiquan Liu and
                  Amin Ul Haq and
                  Yuehui Li and
                  Zoe L. Jiang},
  title        = {Differentially private matrix factorization with sub-linear convergence
                  rate for personalized recommendation},
  journal      = {Inf. Fusion},
  volume       = {126},
  pages        = {103621},
  year         = {2026},
  url          = {https://doi.org/10.1016/j.inffus.2025.103621},
  doi          = {10.1016/J.INFFUS.2025.103621},
}

@inproceedings{RNN1,
  author       = {Di Jin and
                  Rui Wang and
                  Meng Ge and
                  Dongxiao He and
                  Xiang Li and
                  Wei Lin and
                  Weixiong Zhang},
  editor       = {Luc De Raedt},
  title        = {{RAW-GNN:} RAndom Walk Aggregation based Graph Neural Network},
  booktitle    = {Proceedings of the Thirty-First International Joint Conference on
                  Artificial Intelligence, {IJCAI} 2022, Vienna, Austria, 23-29 July
                  2022},
  pages        = {2108--2114},
  publisher    = {ijcai.org},
  year         = {2022},
  url          = {https://doi.org/10.24963/ijcai.2022/293},
  doi          = {10.24963/IJCAI.2022/293},
}

@article{ RNN2,
  title={Recurrent temporal revision graph networks},
  author={Chen, Yizhou and Zeng, Anxiang and Yu, Qingtao and Zhang, Kerui and Yuanpeng, Cao and Wu, Kangle and Huzhang, Guangda and Yu, Han and Zhou, Zhiming},
  journal={Advances in Neural Information Processing Systems},
  volume={36},
  pages={69348--69360},
  year={2023}
}

@article{ RNN3,
  title={Property graph representation learning for node classification},
  author={Li, Shu and Zaidi, Nayyar A and Du, Meijie and Zhou, Zhou and Zhang, Hongfei and Li, Gang},
  journal={Knowledge and Information Systems},
  volume={66},
  number={1},
  pages={237--265},
  year={2024},
  publisher={Springer}
}

@article{MPRZhaohai,
  title={Identification of Key Nodes of Acupoint-Disease Network Based on Motif PageRank Algorithm},
  author={Hai, ZHAO and Jiu, nan, MIAO and Xiao, LIU and Xue, long, YU},
  journal={Journal of Northeastern University (Natural Science)},
  volume={45},
  number={5},
  pages={628-635},
  year={2024}
}

@ARTICLE{MPR,
  author={Zhao, Huan and Xu, Xiaogang and Song, Yangqiu and Lee, Dik Lun and Chen, Zhao and Gao, Han},
  journal={IEEE Transactions on Knowledge and Data Engineering}, 
  title={Ranking Users in Social Networks with Motif-Based PageRank}, 
  year={2021},
  volume={33},
  number={5},
  pages={2179-2192},
  doi={10.1109/TKDE.2019.2953264}}

@article{PageCon,
  author       = {Monica Bianchini and
                  Marco Gori and
                  Franco Scarselli},
  title        = {Inside PageRank},
  journal      = {{ACM} Trans. Internet Techn.},
  volume       = {5},
  number       = {1},
  pages        = {92--128},
  year         = {2005},
  url          = {https://doi.org/10.1145/1052934.1052938},
  doi          = {10.1145/1052934.1052938}
}

@inproceedings{triangular1,
  author       = {Zhongmin Pei and
                  Boren Hu and
                  Zhangkai Luo and
                  Jie Ding},
  title        = {Analysis of node importance of satellite network based on triangular
                  motif},
  booktitle    = {Proceedings of the 2023 International Conference on Communication
                  Network and Machine Learning, {CNML} 2023, Zhengzhou, China, October
                  27-28, 2023},
  pages        = {6--12},
  publisher    = {{ACM}},
  year         = {2023},
  url          = {https://doi.org/10.1145/3640912.3640914},
  doi          = {10.1145/3640912.3640914},
  bibsource    = {dblp computer science bibliography, https://dblp.org}
}

@article{triangular2,
  author       = {Hyeonseong Jeon and
                  Suh{-}Ryung Kim and
                  Dougu Nam and
                  Yun Joo Yoo},
  title        = {Analysis of triangular motifs in protein interaction networks and
                  their implications to protein ages and cancer genes},
  journal      = {Int. J. Data Min. Bioinform.},
  volume       = {19},
  number       = {4},
  pages        = {340--365},
  year         = {2017},
  url          = {https://doi.org/10.1504/IJDMB.2017.10012553},
  doi          = {10.1504/IJDMB.2017.10012553},
  bibsource    = {dblp computer science bibliography, https://dblp.org}
}

@article{cora,
  author       = {Prithviraj Sen and
                  Galileo Namata and
                  Mustafa Bilgic and
                  Lise Getoor and
                  Brian Gallagher and
                  Tina Eliassi{-}Rad},
  title        = {Collective Classification in Network Data},
  journal      = {{AI} Mag.},
  volume       = {29},
  number       = {3},
  pages        = {93--106},
  year         = {2008},
  url          = {https://doi.org/10.1609/aimag.v29i3.2157},
  doi          = {10.1609/AIMAG.V29I3.2157},
  bibsource    = {dblp computer science bibliography, https://dblp.org}
}

@article{amazon,
  author       = {Oleksandr Shchur and
                  Maximilian Mumme and
                  Aleksandar Bojchevski and
                  Stephan G{\"{u}}nnemann},
  title        = {Pitfalls of Graph Neural Network Evaluation},
  journal      = {CoRR},
  volume       = {abs/1811.05868},
  year         = {2018},
  url          = {http://arxiv.org/abs/1811.05868},
  eprinttype    = {arXiv},
  eprint       = {1811.05868},
  bibsource    = {dblp computer science bibliography, https://dblp.org}
}

@article{pubmed,
  title={Query-driven Active Surveying for Collective Classification},
  author={ Namata, Galileo  and  London, Ben  and Namatag, Bert Huang},
  year={2012},
}

@inproceedings{gcn,
  author       = {Thomas N. Kipf and
                  Max Welling},
  title        = {Semi-Supervised Classification with Graph Convolutional Networks},
  booktitle    = {5th International Conference on Learning Representations, {ICLR} 2017,
                  Toulon, France, April 24-26, 2017, Conference Track Proceedings},
  publisher    = {OpenReview.net},
  year         = {2017},
  url          = {https://openreview.net/forum?id=SJU4ayYgl},
}

@inproceedings{gat,
  author       = {Petar Velickovic and
                  Guillem Cucurull and
                  Arantxa Casanova and
                  Adriana Romero and
                  Pietro Li{\`{o}} and
                  Yoshua Bengio},
  title        = {Graph Attention Networks},
  booktitle    = {6th International Conference on Learning Representations, {ICLR} 2018,
                  Vancouver, BC, Canada, April 30 - May 3, 2018, Conference Track Proceedings},
  publisher    = {OpenReview.net},
  year         = {2018},
  url          = {https://openreview.net/forum?id=rJXMpikCZ}
}

@article{
MotifFirst,
author = {R. Milo  and S. Shen-Orr  and S. Itzkovitz  and N. Kashtan  and D. Chklovskii  and U. Alon },
title = {Network Motifs: Simple Building Blocks of Complex Networks},
journal = {Science},
volume = {298},
number = {5594},
pages = {824-827},
year = {2002}
}

@inproceedings{count1,
  author       = {Nesreen K. Ahmed and
                  Jennifer Neville and
                  Ryan A. Rossi and
                  Nick G. Duffield},
  editor       = {Charu C. Aggarwal and
                  Zhi{-}Hua Zhou and
                  Alexander Tuzhilin and
                  Hui Xiong and
                  Xindong Wu},
  title        = {Efficient Graphlet Counting for Large Networks},
  booktitle    = {2015 {IEEE} International Conference on Data Mining, {ICDM} 2015,
                  Atlantic City, NJ, USA, November 14-17, 2015},
  pages        = {1--10},
  publisher    = {{IEEE} Computer Society},
  year         = {2015},
  url          = {https://doi.org/10.1109/ICDM.2015.141},
  doi          = {10.1109/ICDM.2015.141},
}

@inproceedings{count2,
  author       = {Madhav Jha and
                  C. Seshadhri and
                  Ali Pinar},
  editor       = {Aldo Gangemi and
                  Stefano Leonardi and
                  Alessandro Panconesi},
  title        = {Path Sampling: {A} Fast and Provable Method for Estimating 4-Vertex
                  Subgraph Counts},
  booktitle    = {Proceedings of the 24th International Conference on World Wide Web,
                  {WWW} 2015, Florence, Italy, May 18-22, 2015},
  pages        = {495--505},
  publisher    = {{ACM}},
  year         = {2015},
  url          = {https://doi.org/10.1145/2736277.2741101},
  doi          = {10.1145/2736277.2741101},
  bibsource    = {dblp computer science bibliography, https://dblp.org}
}

@inproceedings{count3,
  author       = {Radu Curticapean and
                  Daniel Neuen},
  editor       = {Yossi Azar and
                  Debmalya Panigrahi},
  title        = {Counting Small Induced Subgraphs: Hardness via Fourier Analysis},
  booktitle    = {Proceedings of the 2025 Annual {ACM-SIAM} Symposium on Discrete Algorithms,
                  {SODA} 2025, New Orleans, LA, USA, January 12-15, 2025},
  pages        = {3677--3695},
  publisher    = {{SIAM}},
  year         = {2025},
  url          = {https://doi.org/10.1137/1.9781611978322.122},
  doi          = {10.1137/1.9781611978322.122},
}

@inproceedings{count4,
  author       = {Rong{-}Hua Li and
                  Xiaowei Ye and
                  Fusheng Jin and
                  Yu{-}Ping Wang and
                  Ye Yuan and
                  Guoren Wang},
  editor       = {Guodong Long and
                  Michale Blumestein and
                  Yi Chang and
                  Liane Lewin{-}Eytan and
                  Zi Helen Huang and
                  Elad Yom{-}Tov},
  title        = {Counting Cohesive Subgraphs with Hereditary Properties},
  booktitle    = {Proceedings of the {ACM} on Web Conference 2025, {WWW} 2025, Sydney,
                  NSW, Australia, 28 April 2025- 2 May 2025},
  pages        = {3874--3884},
  publisher    = {{ACM}},
  year         = {2025},
  url          = {https://doi.org/10.1145/3696410.3714730},
  doi          = {10.1145/3696410.3714730},
}

@article{motifweb1,
  author       = {Xin Zheng and
                  Guiling Wang and
                  Guiyue Xu and
                  Jianye Yang and
                  Boyang Han and
                  Jian Yu},
  title        = {A LLM-driven and motif-informed linearizing graph transformer for
                  Web {API} recommendation},
  journal      = {Appl. Soft Comput.},
  volume       = {169},
  pages        = {112547},
  year         = {2025},
  url          = {https://doi.org/10.1016/j.asoc.2024.112547},
  doi          = {10.1016/J.ASOC.2024.112547},
  bibsource    = {dblp computer science bibliography, https://dblp.org}
}

@inproceedings{motifweb2,
  author       = {Rong{-}Hua Li and
                  Xiaowei Ye and
                  Fusheng Jin and
                  Yu{-}Ping Wang and
                  Ye Yuan and
                  Guoren Wang},
  editor       = {Guodong Long and
                  Michale Blumestein and
                  Yi Chang and
                  Liane Lewin{-}Eytan and
                  Zi Helen Huang and
                  Elad Yom{-}Tov},
  title        = {Counting Cohesive Subgraphs with Hereditary Properties},
  booktitle    = {Proceedings of the {ACM} on Web Conference 2025, {WWW} 2025, Sydney,
                  NSW, Australia, 28 April 2025- 2 May 2025},
  pages        = {3874--3884},
  publisher    = {{ACM}},
  year         = {2025},
  url          = {https://doi.org/10.1145/3696410.3714730},
  doi          = {10.1145/3696410.3714730},
  bibsource    = {dblp computer science bibliography, https://dblp.org}
}

@article{shp-gnn,
  author       = {Fei Yang and
                  Huyin Zhang and
                  Shiming Tao and
                  Xiying Fan},
  title        = {Simple hierarchical PageRank graph neural networks},
  journal      = {J. Supercomput.},
  volume       = {80},
  number       = {4},
  pages        = {5509--5539},
  year         = {2024},
  url          = {https://doi.org/10.1007/s11227-023-05666-6},
  doi          = {10.1007/S11227-023-05666-6},
  bibsource    = {dblp computer science bibliography, https://dblp.org}
}

@article{PageRank,
  title={The PageRank Citation Ranking: Bringing Order to the Web},
  author={ Page, Lawrence  and  Brin, Sergey  and  Motwani, Rajeev  and  Winograd, Terry },
  journal={Stanford Digital Libraries Working Paper},
  year={1998},
}

@inproceedings{attention1,
  author       = {Jongmin Park and
                  Seunghoon Han and
                  Soohwan Jeong and
                  Sungsu Lim},
  editor       = {Tat{-}Seng Chua and
                  Chong{-}Wah Ngo and
                  Roy Ka{-}Wei Lee and
                  Ravi Kumar and
                  Hady W. Lauw},
  title        = {Hyperbolic Heterogeneous Graph Attention Networks},
  booktitle    = {Companion Proceedings of the {ACM} on Web Conference 2024, {WWW} 2024,
                  Singapore, Singapore, May 13-17, 2024},
  pages        = {561--564},
  publisher    = {{ACM}},
  year         = {2024},
  url          = {https://doi.org/10.1145/3589335.3651522},
  doi          = {10.1145/3589335.3651522},
  bibsource    = {dblp computer science bibliography, https://dblp.org}
}

@inproceedings{attention3,
  author       = {Huiyuan Chen and
                  Chin{-}Chia Michael Yeh and
                  Fei Wang and
                  Hao Yang},
  editor       = {Fr{\'{e}}d{\'{e}}rique Laforest and
                  Rapha{\"{e}}l Troncy and
                  Elena Simperl and
                  Deepak Agarwal and
                  Aristides Gionis and
                  Ivan Herman and
                  Lionel M{\'{e}}dini},
  title        = {Graph Neural Transport Networks with Non-local Attentions for Recommender
                  Systems},
  booktitle    = {{WWW} '22: The {ACM} Web Conference 2022, Virtual Event, Lyon, France,
                  April 25 - 29, 2022},
  pages        = {1955--1964},
  publisher    = {{ACM}},
  year         = {2022},
  url          = {https://doi.org/10.1145/3485447.3512162},
  doi          = {10.1145/3485447.3512162},
  bibsource    = {dblp computer science bibliography, https://dblp.org}
}

@article{stochastic1,
  author       = {Dong Zhang and
                  Wenlong Feng and
                  Zonghang Wu and
                  Guanyu Li and
                  Bo Ning},
  title        = {{CDRGN-SDE:} Cross-Dimensional Recurrent Graph Network with neural
                  Stochastic Differential Equation for temporal knowledge graph embedding},
  journal      = {Expert Syst. Appl.},
  volume       = {247},
  pages        = {123295},
  year         = {2024},
  url          = {https://doi.org/10.1016/j.eswa.2024.123295},
  doi          = {10.1016/J.ESWA.2024.123295},
  bibsource    = {dblp computer science bibliography, https://dblp.org}
}

@article{stochastic3,
  author       = {Soheila Molaei and
                  Nima Ghanbari Bousejin and
                  Ghadeer O. Ghosheh and
                  Anshul Thakur and
                  Vinod Kumar Chauhan and
                  Tingting Zhu and
                  David A. Clifton},
  title        = {CliqueFluxNet: Unveiling {EHR} Insights with Stochastic Edge Fluxing
                  and Maximal Clique Utilisation Using Graph Neural Networks},
  journal      = {J. Heal. Informatics Res.},
  volume       = {8},
  number       = {3},
  pages        = {555--575},
  year         = {2024},
  url          = {https://doi.org/10.1007/s41666-024-00169-2},
  doi          = {10.1007/S41666-024-00169-2},
  bibsource    = {dblp computer science bibliography, https://dblp.org}
}

@inproceedings{graph11,
  author       = {Qi Chen and
                  Zhiqiong Wang and
                  Jiaxin Li and
                  Jinying Tao and
                  Junchang Xin},
  title        = {{TSTAI:} {A} Time-varying Brain Effective Connectivity Network Construction
                  Method Combining with Brain Active Information},
  booktitle    = {Proceedings of the Thirty-Fourth International Joint Conference on
                  Artificial Intelligence, {IJCAI} 2025, Montreal, Canada, August 16-22,
                  2025},
  pages        = {7347--7355},
  publisher    = {ijcai.org},
  year         = {2025},
  url          = {https://doi.org/10.24963/ijcai.2025/817},
  doi          = {10.24963/IJCAI.2025/817},
}

@article{graph3,
  author       = {Hanchen Wang and
                  Ying Zhang and
                  Wenjie Zhang},
  title        = {Machine Learning for Graph Data Management and Query Processing},
  journal      = {Proc. {VLDB} Endow.},
  volume       = {18},
  number       = {12},
  pages        = {5499--5503},
  year         = {2025},
  url          = {https://www.vldb.org/pvldb/vol18/p5499-zhang.pdf},
  bibsource    = {dblp computer science bibliography, https://dblp.org}
}

@inproceedings{graph1,
  author       = {Yonghao Liu and
                  Mengyu Li and
                  Fausto Giunchiglia and
                  Lan Huang and
                  Ximing Li and
                  Xiaoyue Feng and
                  Renchu Guan},
  editor       = {Guodong Long and
                  Michale Blumestein and
                  Yi Chang and
                  Liane Lewin{-}Eytan and
                  Zi Helen Huang and
                  Elad Yom{-}Tov},
  title        = {Dual-level Mixup for Graph Few-shot Learning with Fewer Tasks},
  booktitle    = {Proceedings of the {ACM} on Web Conference 2025, {WWW} 2025, Sydney,
                  NSW, Australia, 28 April 2025- 2 May 2025},
  pages        = {2646--2656},
  publisher    = {{ACM}},
  year         = {2025},
  url          = {https://doi.org/10.1145/3696410.3714905},
  doi          = {10.1145/3696410.3714905},
}

@inproceedings{graph5,
  author       = {Mingyang Zhou and
                  Gang Liu and
                  Kezhong Lu and
                  Hao Liao and
                  Rui Mao},
  editor       = {Guodong Long and
                  Michale Blumestein and
                  Yi Chang and
                  Liane Lewin{-}Eytan and
                  Zi Helen Huang and
                  Elad Yom{-}Tov},
  title        = {Highly-efficient Minimization of Network Connectivity in Large-scale
                  Graphs},
  booktitle    = {Proceedings of the {ACM} on Web Conference 2025, {WWW} 2025, Sydney,
                  NSW, Australia, 28 April 2025- 2 May 2025},
  pages        = {2530--2539},
  publisher    = {{ACM}},
  year         = {2025},
  url          = {https://doi.org/10.1145/3696410.3714806},
  doi          = {10.1145/3696410.3714806},
}

@ARTICLE{Zhao,
  author={Zhao, Chengjie and Wang, Jun and Peng, Qihang and Huang, Wei and Chen, Xiaonan and Zhao, Zexue and Le-Ngoc, Tho},
  journal={IEEE Transactions on Cognitive Communications and Networking}, 
  title={Deep Learning Based Transceiver Design for Additive Non-Gaussian Impulsive Noise Channels}, 
  year={2025},
  volume={},
  number={},
  pages={1-1},
  doi={10.1109/TCCN.2025.3547726}}

@inproceedings{wa,
  author       = {Badih Ghazi and
                  Ravi Kumar and
                  Pasin Manurangsi},
  editor       = {Tat{-}Seng Chua and
                  Chong{-}Wah Ngo and
                  Roy Ka{-}Wei Lee and
                  Ravi Kumar and
                  Hady W. Lauw},
  title        = {Privacy in Web Advertising: Analytics and Modeling},
  booktitle    = {Companion Proceedings of the {ACM} on Web Conference 2024, {WWW} 2024,
                  Singapore, Singapore, May 13-17, 2024},
  pages        = {1288--1289},
  publisher    = {{ACM}},
  year         = {2024},
  url          = {https://doi.org/10.1145/3589335.3641252},
  doi          = {10.1145/3589335.3641252}
}

@inproceedings{sna,
  author       = {Yurui Lai and
                  Xiaoyang Lin and
                  Renchi Yang and
                  Hongtao Wang},
  editor       = {Ricardo Baeza{-}Yates and
                  Francesco Bonchi},
  title        = {Efficient Topology-aware Data Augmentation for High-Degree Graph Neural
                  Networks},
  booktitle    = {Proceedings of the 30th {ACM} {SIGKDD} Conference on Knowledge Discovery
                  and Data Mining, {KDD} 2024, Barcelona, Spain, August 25-29, 2024},
  pages        = {1463--1473},
  publisher    = {{ACM}},
  year         = {2024},
  url          = {https://doi.org/10.1145/3637528.3671765},
  doi          = {10.1145/3637528.3671765},
  bibsource    = {dblp computer science bibliography, https://dblp.org}
}

@article{sl1,
  author       = {MoonJeong Park and
                  Sunghyun Choi and
                  Jaeseung Heo and
                  Eunhyeok Park and
                  Dongwoo Kim},
  title        = {The Oversmoothing Fallacy: {A} Misguided Narrative in {GNN} Research},
  journal      = {CoRR},
  volume       = {abs/2506.04653},
  year         = {2025},
  url          = {https://doi.org/10.48550/arXiv.2506.04653},
  doi          = {10.48550/ARXIV.2506.04653},
  eprinttype    = {arXiv},
  eprint       = {2506.04653},
  bibsource    = {dblp computer science bibliography, https://dblp.org}
}

@inproceedings{sl3,
author = {Zhao, Weichen and Wang, Chenguang and Wang, Xinyan and Han, Congying and Guo, Tiande and Yu, Tianshu},
title = {Understanding Oversmoothing in Diffusion-Based GNNs From the Perspective of Operator Semigroup Theory},
year = {2025},
isbn = {9798400712456},
publisher = {Association for Computing Machinery},
address = {New York, NY, USA},
url = {https://doi.org/10.1145/3690624.3709324},
doi = {10.1145/3690624.3709324},
booktitle = {Proceedings of the 31st ACM SIGKDD Conference on Knowledge Discovery and Data Mining V.1},
pages = {2043–2054},
numpages = {12},
location = {Toronto ON, Canada},
series = {KDD '25}
}

@article{ol1,
  author       = {Nicolas Keriven},
  title        = {Backward Oversmoothing: why is it hard to train deep Graph Neural
                  Networks?},
  journal      = {CoRR},
  volume       = {abs/2505.16736},
  year         = {2025},
  url          = {https://doi.org/10.48550/arXiv.2505.16736},
  doi          = {10.48550/ARXIV.2505.16736},
  eprinttype    = {arXiv},
  eprint       = {2505.16736},
  bibsource    = {dblp computer science bibliography, https://dblp.org}
}

@InProceedings{ol3,
  title = 	 {Graph Neural Networks (with Proper Weights) Can Escape Oversmoothing},
  author =       {Zhuo, Zhijian and Wang, Yifei and Ma, Jinwen and Wang, Yisen},
  booktitle = 	 {Proceedings of the 16th Asian Conference on Machine Learning},
  pages = 	 {17--32},
  year = 	 {2025},
  editor = 	 {Nguyen, Vu and Lin, Hsuan-Tien},
  volume = 	 {260},
  series = 	 {Proceedings of Machine Learning Research},
  month = 	 {05--08 Dec},
  publisher =    {PMLR},
  url = 	 {https://proceedings.mlr.press/v260/zhuo25a.html}
}

@article{ol5,
  author       = {Dimitrios Kelesis and
                  Dimitris Fotakis and
                  Georgios Paliouras},
  title        = {Reducing oversmoothing through informed weight initialization in graph
                  neural networks},
  journal      = {Appl. Intell.},
  volume       = {55},
  number       = {7},
  pages        = {632},
  year         = {2025},
  url          = {https://doi.org/10.1007/s10489-025-06426-0},
  doi          = {10.1007/S10489-025-06426-0},
  bibsource    = {dblp computer science bibliography, https://dblp.org}
}

@inproceedings{gin,
  author       = {Keyulu Xu and
                  Weihua Hu and
                  Jure Leskovec and
                  Stefanie Jegelka},
  title        = {How Powerful are Graph Neural Networks?},
  booktitle    = {7th International Conference on Learning Representations, {ICLR} 2019,
                  New Orleans, LA, USA, May 6-9, 2019},
  publisher    = {OpenReview.net},
  year         = {2019},
  url          = {https://openreview.net/forum?id=ryGs6iA5Km},
  bibsource    = {dblp computer science bibliography, https://dblp.org}
}

@article{VGAE,
  author       = {Thomas N. Kipf and
                  Max Welling},
  title        = {Variational Graph Auto-Encoders},
  journal      = {CoRR},
  volume       = {abs/1611.07308},
  year         = {2016},
  url          = {http://arxiv.org/abs/1611.07308},
  eprinttype    = {arXiv},
  eprint       = {1611.07308},
  bibsource    = {dblp computer science bibliography, https://dblp.org}
}

@inproceedings{HDGL,
  author       = {Abhishek Dalvi and
                  Vasant G. Honavar},
  editor       = {Wolfgang Nejdl and
                  S{\"{o}}ren Auer and
                  Meeyoung Cha and
                  Marie{-}Francine Moens and
                  Marc Najork},
  title        = {Hyperdimensional Representation Learning for Node Classification and
                  Link Prediction},
  booktitle    = {Proceedings of the Eighteenth {ACM} International Conference on Web
                  Search and Data Mining, {WSDM} 2025, Hannover, Germany, March 10-14,
                  2025},
  pages        = {88--97},
  publisher    = {{ACM}},
  year         = {2025},
  url          = {https://doi.org/10.1145/3701551.3703492},
  doi          = {10.1145/3701551.3703492},
  bibsource    = {dblp computer science bibliography, https://dblp.org}
}

@article{PR_add,
  author       = {Haoyu Liu and
                  Siqiang Luo},
  title        = {{BIRD:} Efficient Approximation of Bidirectional Hidden Personalized
                  PageRank},
  journal      = {Proc. {VLDB} Endow.},
  volume       = {17},
  number       = {9},
  pages        = {2255--2268},
  year         = {2024},
  url          = {https://www.vldb.org/pvldb/vol17/p2255-liu.pdf},
  doi          = {10.14778/3665844.3665855},
  bibsource    = {dblp computer science bibliography, https://dblp.org}
}

@article{add00001,
  author       = {Xiaoyang Lin and
                  Runhao Jiang and
                  Renchi Yang},
  title        = {Effective Clustering for Large Multi-Relational Graphs},
  journal      = {Proc. {ACM} Manag. Data},
  volume       = {3},
  number       = {6},
  pages        = {1--28},
  year         = {2025},
  url          = {https://doi.org/10.1145/3769784},
  doi          = {10.1145/3769784},
  bibsource    = {dblp computer science bibliography, https://dblp.org}
}

@inproceedings{add00002,
  author       = {Haoran Zheng and
                  Jieming Shi and
                  Renchi Yang},
  editor       = {Toby Walsh and
                  Julie Shah and
                  Zico Kolter},
  title        = {GraSP: Simple Yet Effective Graph Similarity Predictions},
  booktitle    = {AAAI-25, Sponsored by the Association for the Advancement of Artificial
                  Intelligence, February 25 - March 4, 2025, Philadelphia, PA, {USA}},
  pages        = {22884--22892},
  publisher    = {{AAAI} Press},
  year         = {2025},
  url          = {https://doi.org/10.1609/aaai.v39i21.34450},
  doi          = {10.1609/AAAI.V39I21.34450},
  bibsource    = {dblp computer science bibliography, https://dblp.org}
}

\newpage
\appendix

\section*{Appendix}
\section{The Motif-Based Adjacency Matrices Calculation of the Seven Considered Motifs.}
\label{appendix7motifs}
In this section, the motif-based adjacency matrices calculation process of seven kinds of motifs are represented. For a graph $G$, whose adjacency matrix is marked as $\boldsymbol{A}$, let $\boldsymbol{U}$ ($\boldsymbol{B}$, respectively) be the adjacency matrix of the unidirectional (bidirectional, respectively) links of $G$. Then the motif-based adjacency matrices of graph $G$ with respect to $M_1$ to $M_7$ is calculated as \autoref{table-m17}. 

\begin{table}[!h]
\begin{threeparttable}
\caption{The Motif-Based Adjacency Matrices Calculation. \label{table-m17}}
\renewcommand{\arraystretch}{1.05}
\begin{tabular}{lll}\hline
Motif & Intermediate   Matrix & $A_{M_i}$ \\ \hline
$M_1$ & $\boldsymbol{\zeta} =\left( \boldsymbol{U}\cdot \boldsymbol{U} \right)\odot {{\boldsymbol{U}}^{T}}$         & $\boldsymbol{\zeta} +{{\boldsymbol{\zeta} }^{T}}$ \\
$M_2$      & $\boldsymbol{\zeta} =\left( \boldsymbol{B}\cdot \boldsymbol{U} \right)\odot {{\boldsymbol{U}}^{T}}+\left( \boldsymbol{U}\cdot \boldsymbol{B} \right)\odot {{\boldsymbol{U}}^{T}}+\left( \boldsymbol{U}\cdot \boldsymbol{U} \right)\odot \boldsymbol{B}$                      &$\boldsymbol{\zeta} +{{\boldsymbol{\zeta} }^{T}}$  \\
$M_3$      & $\boldsymbol{\zeta} =\left( \boldsymbol{B}\cdot \boldsymbol{B} \right)\odot \boldsymbol{U}+\left( \boldsymbol{B}\cdot \boldsymbol{U} \right)\odot \boldsymbol{B}+\left( \boldsymbol{U}\cdot \boldsymbol{B} \right)\odot \boldsymbol{B}$                      & $\boldsymbol{\zeta} +{{\boldsymbol{\zeta} }^{T}}$ \\
$M_4$      & $\boldsymbol{\zeta} =\left( \boldsymbol{B}\cdot \boldsymbol{B} \right)\odot \boldsymbol{B}$                      & $\boldsymbol{\zeta}$ \\
$M_5$      & $\boldsymbol{\zeta} =\left( \boldsymbol{U}\cdot \boldsymbol{U} \right)\odot \boldsymbol{U}+\left( \boldsymbol{U}\cdot {{\boldsymbol{U}}^{T}} \right)\odot \boldsymbol{U}+\left( {{\boldsymbol{U}}^{T}}\cdot \boldsymbol{U} \right)\odot \boldsymbol{U}$                      & $\boldsymbol{\zeta} +{{\boldsymbol{\zeta} }^{T}}$ \\
$M_6$      & $\boldsymbol{\zeta} =\left( \boldsymbol{U}\cdot \boldsymbol{B} \right)\odot \boldsymbol{U}+\left( \boldsymbol{B}\cdot {{\boldsymbol{U}}^{T}} \right)\odot {{\boldsymbol{U}}^{T}}+\left( {{\boldsymbol{U}}^{T}}\cdot \boldsymbol{U} \right)\odot \boldsymbol{B}$                      & $\boldsymbol{\zeta}$ \\
$M_7$      & $\boldsymbol{\zeta} =\left( {{\boldsymbol{U}}^{T}}\cdot \boldsymbol{B} \right)\odot {{\boldsymbol{U}}^{T}}+\left( \boldsymbol{B}\cdot \boldsymbol{U} \right)\odot \boldsymbol{U}+\left( \boldsymbol{U}\cdot {{\boldsymbol{U}}^{T}} \right)\odot \boldsymbol{B}$                      & $\boldsymbol{\zeta}$	\\ \hline
\end{tabular}
\begin{tablenotes} 
\footnotesize
\item[] The table is grabbed from \cite{MotifDef, MPR}. $\odot$ denotes the Hadamard (entry-wise) product.
\end{tablenotes}
\end{threeparttable}
\end{table}
\section{The Details of the Datasets. }\label{appendixdataset}
Cora \cite{cora} and PubMed \cite{pubmed} are famous and widely applied citation graphs, where nodes represent papers and the edges represent citations between them. Amazon Computers and Amazon Photo \cite{amazon} are both segments of the Amazon co-purchase graph, where nodes represent goods, edges indicate that two goods are frequently bought together, node features are bag-of-words encoded product reviews, and class labels are given by the product category. The statistics information of these networks are as shown in \autoref{tabledataset}. 

\begin{table}[!h]
\caption{The Statistics of Datasets. \label{tabledataset}}
\renewcommand{\arraystretch}{1.05}
\begin{tabular}{lcccc}
\hline
                 & Classes & Features & Nodes & Edges  \\ \hline
Cora             & 7       & 1433     & 2845  & 5069   \\
PubMed           & 3       & 500      & 19717 & 44324  \\
Amazon Computers & 10      & 767      & 13381 & 245778 \\
Amazon Photo     & 8       & 745      & 7487  & 119043 \\ \hline
\end{tabular}
\end{table}

\section{Analysis of the stability of GCN-MPPR }
\label{appendixbox}
\begin{figure}[!h] 
    \centering  
    \begin{subfigure}[b]{0.20\textwidth}  
        \centering
        \includegraphics[width=\textwidth]{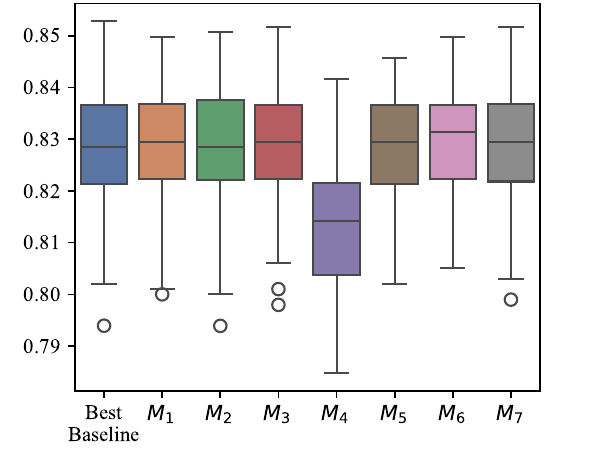}  
        \caption{Cora}  
    \end{subfigure}
    \begin{subfigure}[b]{0.20\textwidth}
        \centering
        \includegraphics[width=\textwidth]{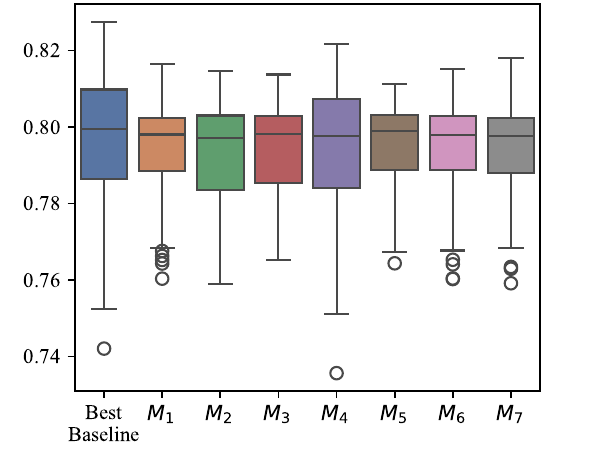}
        \caption{PubMed}
    \end{subfigure}
    \\
    \begin{subfigure}[b]{0.20\textwidth}
        \centering
        \includegraphics[width=\textwidth]{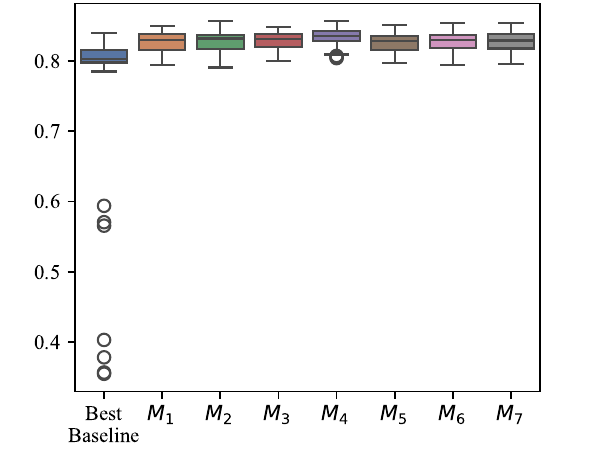}
        \caption{Amazon Computers}
    \end{subfigure}
    \begin{subfigure}[b]{0.20\textwidth}
        \centering
        \includegraphics[width=\textwidth]{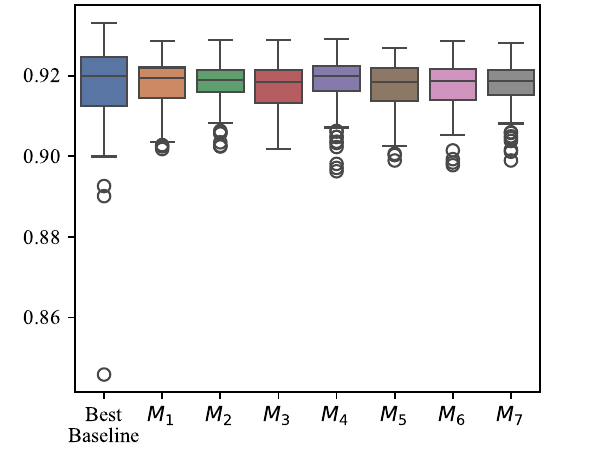}
        \caption{Amazon Photo}
    \end{subfigure}
    \caption{The Box Figures of a Hundred Runs on Various Datasets. }  
    \label{fig4boxes}  
\end{figure}

In order to further verify the stability of GCN-MPPR, the classification accuracy of each run on multiple networks is collected and presented in \autoref{fig4boxes}. In each box figures, the box on the far left is the best performer of each network, while the other 7 boxes represent the results of GCN-MPPR with $M_1$ to $M_7$. Although in some cases the top limit of accuracy achieved by the best baselines is higher, in almost all cases the lower limit of GCN-MPPR has been significantly improved. Meanwhile. The box of GCN-MPPR is comparatively shorter. As a result, together with \autoref{table-acc}, it is confirmed that GCN-MPPR not only improves the accuracy but also enhances the stability.  

\section{The Details of the GCN-MPPR Implementation on DGCRL. }
\label{appendixDGCRL}

In DGCRL, the model of GCN is applied. In order to implement GCN-MPPR upon DGCRL, we multiply the calculated MPPR to the output of the encoder on DGCRL, thereby making the information of each node transfer deeper and further without adding layers. And all of other technical details are hold on still without any modification. As for the experiment, the settings are list in \autoref{table-DGCRL}.  

\begin{table}[!h]
\caption{Implementation Parameter Details of GCN-MPPR on DGCRL. \label{table-DGCRL}}
\setlength{\tabcolsep}{1mm}
\renewcommand{\arraystretch}{1.05}
\begin{tabular}{llll} \hline
Parameter                   & Value & Parameter                   & Value     \\ \hline
Train/Val/Test              & 2:4:4 &  Hidden Dimonsion of Layer   & 128     \\
Training Epochs             & 500    & Out Dimonsion of Encoder    & 16   \\
Learning Rate               & 0.01  &  Hidden Dimonsion of Decoder & 128  \\
$p_{r, 1}$, $p_{r, 2}$, $p_{m, 1}$, $p_{m, 2}$             & 0.1  & Negative Sampling Ratio     & 1.0  \\
\hline
\end{tabular}
\end{table}
\end{document}